\documentclass[12pt,a4paper]{report}
\usepackage[utf8]{inputenc}
\usepackage[T1]{fontenc}
\usepackage{graphicx}
\usepackage{amsmath}
\usepackage{amssymb}
\usepackage{geometry}
\usepackage{setspace}
\usepackage{hyperref}
\usepackage{times} 
\usepackage{titlesec}
\usepackage{tocloft}
\usepackage{multirow}
\usepackage{float}
\usepackage{array} 
\usepackage{longtable}
\usepackage{caption}

\renewcommand{\arraystretch}{1.2}
\hypersetup{
    colorlinks=true,
    linkcolor=black,
    filecolor=magenta,
    urlcolor=black,
}

\titleformat{\chapter}[hang]
{\normalfont\LARGE\bfseries}
{\thechapter.}{0.5em}{}
\titlespacing*{\chapter}{0pt}{30pt}{20pt}

\begin{document}

\begin{titlepage}
    \begin{center}
        \vspace*{1cm}
        
        {\large\textbf{B.Sc. (Business Analytics) Dissertation}\par}
        \vspace{1.5cm}
        
        {\Large\textbf{Enhancing Sports Strategy with Video Analytics and Data Mining}\par}
        \vspace{0.5cm}

        \vspace{3cm}
        {\large By\par}
        {\large\bfseries Chen Jia Wei\par}
        \vspace{5cm}
        
        {\large Department of Information Systems and Analytics\\
        School of Computing\\
        National University of Singapore\par}
        \vspace{1cm}
        
        {\large 2024/2025\par}
    \end{center}
\end{titlepage}

\begin{titlepage}
    \begin{center}
        \vspace*{1cm}
        
        {\large\textbf{B.Sc. (Business Analytics) Dissertation}\par}
        \vspace{1.5cm}
        
        {\Large\textbf{Enhancing Sports Strategy with Video Analytics and Data Mining: Automated Video-Based Analytics Framework for Tennis Doubles}\par}
        \vspace{1.5cm}
        
        \vspace{1cm}
        {\large By\par}
        {\large\bfseries Chen Jia Wei\par}
        \vspace{2cm}
        
        {\large Department of Information Systems and Analytics\\
        School of Computing\\
        National University of Singapore\par}
        \vspace{1cm}
        
        {\large 2024/2025\par}
    \end{center}
    
    \vfill
    
    \begin{flushleft}
        {\large\textbf{Project ID:} H403030\par}
        {\large\textbf{Advisor:} Dr. Jiang Kan\par}
        \vspace{0.3cm}
        {\large\textbf{Deliverables:}\par}
        {\normalsize  Report: 1 Volume\par}
        {\normalsize  Software: \href{https://github.com/alvinjiang1/tennis-annotation-tool}{\texttt{Tennis Annotation Tool}} \href{https://github.com/jiaawe/tennis-prediction}{\texttt{\& Prediction Models}}\par}
        {\normalsize  Annotation Tool Guide:  \href{https://youtu.be/nXhNrud8wdU}{\texttt{Demo}}\par}
    \end{flushleft}
\end{titlepage}

\chapter*{Abstract}
\addcontentsline{toc}{chapter}{Abstract}
My final year project introduces a comprehensive video-based analytics framework specifically tailored for tennis doubles, addressing a significant gap in automated analysis tools for this strategically complex sport. The framework features a standardised annotation methodology designed explicitly for doubles tennis, encompassing critical elements such as player positioning, shot types, court formations, and outcomes.

In addition, a specialised tennis annotation tool was developed, addressing functional and non-functional requirements unique to tennis labelling. Advanced machine learning techniques, including GroundingDINO for precise player localisation through descriptive phrase grounding and YOLO-Pose for robust pose estimation were integrated. These methods significantly reduce manual annotation efforts while enhancing the consistency and quality of collected data.

Experimental results show superior effectiveness of Convolutional Neural Networks (CNN) based models, utilising transfer learning, in automatically predicting essential match details such as shot type, player positioning, and formations. These models notably outperformed pose-based approaches, effectively capturing complex visual and contextual features crucial for doubles tennis analytics.

The integrated annotation tool successfully merges advanced analytical capabilities with the strategic complexities of tennis doubles, creating a solid foundation for future research, strategic modelling, performance analysis, and automated tactical insights.

\vspace{0.3cm}
\noindent \textbf{Subject Descriptors:}
\begin{itemize}
    \item I.2: ARTIFICIAL INTELLIGENCE
    \item I.4: IMAGE PROCESSING AND COMPUTER VISION
    \item D.2: SOFTWARE ENGINEERING
    \item I.5: PATTERN RECOGNITION
    \item H.1.1: Systems and Information Theory
\end{itemize}

\vspace{0.3cm}
\noindent \textbf{Keywords:} Sports Analytics, Machine Learning, Video Annotation, Pose Estimation, Shot Classification, Transfer Learning, Visual Grounding, Software Engineering

\vspace{0.3cm}
\noindent \textbf{Implementation Software:} React, Vite, DaisyUI, TailwindCSS, Flask, FFmpeg, PyTorch

\chapter*{Acknowledgments}
\addcontentsline{toc}{chapter}{Acknowledgments}
In this section, I would like to show my appreciation and give credit to my professor, teammates and everyone who has contributed and supported my final year project in the domain of sports analytics for tennis doubles.

First and foremost, I would like to extend my sincere gratitude to Professor Jiang Kan for his invaluable guidance, insightful feedback, and unwavering support throughout the duration of this research. His expertise and vision were instrumental in defining the project's scope and shaping our methodological approach. Throughout the year-long journey, he has never failed to encourage me to think critically and explore novel yet practical ways in my approach to this project.

I also extend my heartfelt appreciation to Zhaoyu Liu, whose constructive critiques, technical suggestions, and commitment to our weekly progress updates significantly enhanced the quality of this work. His attention to methodological rigor and research integrity was instrumental in navigating the challenges we encountered throughout this project.

I would also like to acknowledge my teammates who contributed to various aspects of this project. Special thanks to Alvin Jiang Min Jun for his close collaboration on key components of the application, particularly his research efforts in identifying the GroundingDINO model, which was ultimately implemented in our final labelling tool. He also helped in building several components of the frontend and backend for the tennis annotation tool. His partnership was vital in addressing several technical challenges we faced.

Next, to Chen Guan Zhou, my appreciation for his parallel work in the tennis doubles domain. Though focusing on downstream applications, his research complemented our efforts and provided valuable context that enhanced our understanding of the practical implications of our work. Alongside Alvin, he also worked together with me to label tennis doubles videos to be used for downstream model training.

This project represents not just my individual effort but the collective intelligence and dedication of a team committed to advancing sports analytics in the underexplored domains, such as tennis doubles. The support of everyone mentioned above was crucial in developing this integrated framework for automated doubles tennis analysis.

\clearpage

\tableofcontents
\clearpage

\listoffigures
\clearpage

\listoftables
\clearpage

\chapter{Introduction}
\section{Background Information}
Sports analytics is the process of deriving actionable insights from athletic performances through the usage of data science and statistical analysis. This fast-evolving field has transformed the way many sports are played and strategised, such as match outcome predictions, performance optimisation, and strategic decision-making. While sports analytics has made significant strides in team sports such as basketball and soccer, as well as individual sports in tennis singles or badminton, certain domains remain relatively under-explored due to its' strategic complexity and uniqueness in the way they are played.

Tennis doubles is one such domain that is under-explored. The applications of robust tennis doubles analytics extend across multiple stakeholders in the sport: player \& coaching development, strategic outcome modelling, performance tracking and other professional analysis.

\section{Problem Statement}
The dynamic and strategic nature of tennis doubles presents unique analytical challenges that set it apart from singles play. While professional singles tennis has seen significant advancement in match analytics, doubles tennis remains relatively underexplored in terms of automated analysis tools. The complexity of having four players on court, intricate positioning strategies, and rapid partner-to-partner interactions makes doubles tennis particularly challenging for traditional analysis methods. 

As of now, coaches and players often rely on manual video analysis, which is both time-consuming and potentially subjective, limiting the depth and scale of tactical insights that can be derived from match footage.

Furthermore, the lack of standardised doubles-specific annotation frameworks means that even when data is collected, it often lacks the structure and comprehensiveness needed for advanced analytical applications. This gap between the strategic complexity of doubles tennis and the tools available to analyse it creates a significant opportunity for innovation in sports analytics technology.

\section{Project Objectives}
This project aims to develop an integrated framework for automated doubles tennis analysis to address the challenges highlighted above. The primary objectives are:

\begin{itemize}
\item \textbf{Development of a Standardised Video Annotation Framework for Tennis Doubles}
\begin{itemize}
\item Define comprehensive event types and data points specific to doubles tennis, including player positions, shot types, formations, and court events
\item Establish systematic data collection methodologies that balance analytical depth with practical implementation
\item Create initial annotated datasets to support model development and baseline analysis
\end{itemize}

\item \textbf{Implementation of an Intuitive Tennis Labelling Tool}
\begin{itemize}
\item Design and develop a user-friendly application tailored specifically for doubles tennis scenarios
\item Ensure the tool follows software engineering best practices for maintainability and extensibility, while optimising the interface for efficiency to reduce the time burden of manual annotation
\item Structure the application to seamlessly incorporate automated analysis components as they are developed
\end{itemize}

\item \textbf{Automation of Data Collection and Analysis Processes}
\begin{itemize}
\item Implement player detection and tracking systems to automatically locate and follow all four players throughout matches based on their given descriptions
\item Develop pose estimation capabilities to capture players' body positions without manual annotation
\item Create classification models to automatically identify shot types, formations, and other key events defined in the framework
\item Integrate these automated components into the labelling tool to progressively reduce manual effort
\end{itemize}
\end{itemize}

By achieving these objectives, this project aims to provide a tool and framework to bridge the gap between the complex tactical nature of doubles tennis and the analytical tools available to coaches, players, and researchers. The resulting framework will facilitate a more comprehensive analysis of doubles matches and enable advanced applications such as strategic modeling, performance optimisation, and automated tactical insights.

\section{Project Scope}
The project covers the creation of a (1) standardised annotation taxonomy specific to tennis doubles (shot types, court positions, formations, etc.), (2) development of an intuitive and scalable video labelling application supporting manual and semi-annotation capabilities, (3) implementation of computer vision models for player detection, tracking, and pose estimation, (4) design of algorithms for basic shot classification and (5) annotation, testing and labelling of both amateur and professional tennis doubles match footage to make available an initial tennis doubles dataset.

\section{Project Significance}
This project addresses a popular yet overlooked area, tennis doubles, in sports analytics. Despite its popularity and intricate strategies, tennis doubles has seen limited analytical exploration. The significance of this work is evident in multiple areas.

\textbf{Innovative Analytical Methods:} By developing specialised approaches for multi-player tracking, pose analysis, and shot type classification, this project pushes the boundaries of sports data analysis in complex environments. These innovative methods are adaptable and could be beneficially applied to other team sports that experience similar challenges.

\textbf{Foundational Data Framework:} Establishing a standardised approach to data collection enables advanced analyses such as strategic modelling through Markov Decision Processes (MDPs), tactical pattern recognition using machine learning, and detailed performance evaluation. Our proposed data infrastructure lays the groundwork for an expanding ecosystem of analytics specifically tailored for tennis doubles in the future.

\textbf{Linking Practice and Research:} The project creates valuable connections between practical coaching applications and academic research fields, including sports science, computer vision, and decision theory. By offering structured and reliable data about tennis doubles play, researchers gain a powerful avenue and resource to test their theories and build practical models that coaches can directly utilise, made possible by the project's comprehensive labelling framework, analytical tools, and datasets.

\chapter{Existing Work}
\section{Sports Analytics}
Recent works in sports analytics have explored a wide range of challenges, including strategy modeling using probabilistic reasoning and deep learning (Dong et al., 2023; Liu et al., 2023a; Liu et al., 2023b; Liu et al., 2024a; Liu et al., 2024b; Liu et al., 2024c; Hundal et al., 2024), injury prediction (Liu et al., 2023c), fine-grained event detection in sports videos (Liu et al., 2025a), and specific tasks such as court detection and ball tracking in broadcast footage (Jiang et al., 2020; Jiang et al., 2023; Jiang et al., 2024). These efforts demonstrate the growing integration of AI, data mining, and formal methods in advancing sports performance analysis and tactical understanding (Liu et al., 2025b).

\section{Video Annotation Tools for Sports Analytics}
Over the past several years, several video annotation tools have been developed for sports analytics, ranging from general-purpose labeling software to specialised sports analysis platforms. These tools vary in terms of automation, usability, and applicability to tennis doubles analysis.

\textbf{Existing Annotation Tools}
\textbf{Open-Source Annotation Platforms:} Several open-source tools provide video labelling functionalities for computer vision and machine learning applications. One widely used platform is the Computer Vision Annotation Tool (CVAT), developed by Intel, which supports image and video annotation, including object tracking and frame interpolation (Intel, 2025). Another tool, the VGG Image Annotator (VIA), is a lightweight annotation tool developed by the Visual Geometry Group at Oxford, designed for manual and semi-automated labelling of images and videos (Dutta \& Zisserman, 2019). Additionally, Label Studio has gained popularity as an open-source data labelling tool that supports video, image, and text annotations, enabling the integration of Artificial Intelligence (AI) assisted labelling workflows (HumanSignal, 2025).

For sports-specific analysis, Kinovea is a free and open-source tool widely used by coaches for motion analysis, slow-motion replay, and basic annotation (Kinovea, 2023). While useful for tagging and analysing player movements, it lacks the scalability and structure required for automated dataset creation in machine learning applications, especially true for Tennis doubles' context.

\textbf{Commercialised and Specialised Tools:} Several paid platforms provide more advanced annotation capabilities, often integrating AI-assisted features. Encord and V7 Labs offer cloud-based annotation tools with AI-assisted tracking and segmentation (Encord, 2025; V7 Labs, 2025). These tools streamline the annotation process but require cloud-based data storage and subscriptions. In the sports industry, Dartfish and Hudl Sportscode are commonly used for performance analysis. Dartfish allows users to tag and categorise video clips, add custom labels, and generate performance reports (Dartfish, 2025). Similarly, SwingVision, an AI-powered tennis analysis tool, provides automated shot tracking, ball speed estimation, and real-time feedback (SwingVision, 2025). However, these applications primarily focus on coaching and player improvement rather than dataset generation for doubles-specific machine learning models.

\textbf{Limitations in Existing Annotation Tools:}
Most existing annotation tools are optimised for general-purpose video labeling or single-player sports analysis. Doubles tennis introduces additional complexity, such as tracking four players, capturing tactical interactions, and identifying coordinated movements, which general annotation platforms do not inherently support. For instance, CVAT and VIA allow manual annotation of players and the ball, but they do not provide a structured way to link player roles (i.e., net player, baseline player) or automate the tracking of team formations. Similarly, Kinovea supports basic movement tracking but does not facilitate structured, scalable annotation of doubles matches.

A recent study by Wang, Lai, Huang \& Lin (2024) demonstrated the challenges of tennis video annotation. They manually labelled player key points in video frames and categorised actions (forehand, backhand, serve) using a generic object annotation tool, highlighting the time-intensive nature of manual labelling. A dedicated doubles tennis annotation tool could alleviate these challenges by automating player tracking, capturing interactions between partners, and standardising event labels for machine learning applications.

\section{Potential Research Benefits from a Tennis Doubles Dataset}
Recent years have seen numerous studies in tennis analytics that rely on video-derived data. These span pose estimation, player tracking, shot classification, strategic analysis, and predictive modelling. A common thread is that better labeled data leads to better models. Here we review key works in these areas, noting their methodologies and how a dedicated doubles labelling framework could accelerate progress.

\textbf{Pose Estimation and Player Tracking:}
Pose estimation is crucial for analysing player biomechanics and movement patterns in tennis. Wang et al. (2024) developed a dataset of tennis player poses by manually labelling key points on thousands of video frames, facilitating pose estimation model training. Similarly, AlShami, Boult, \& Kalita (2023) introduced Pose2Trajectory, a deep learning system that predicts player movement trajectories based on pose data extracted from video sequences. Their work demonstrated the importance of high-quality labelled pose data for accurate motion prediction. However, both studies relied on manual labelling, underscoring the need for automated annotation tools in tennis.

Tracking all four players and the ball in doubles tennis presents additional challenges. Unlike singles, where players independently decide on their positioning and play, doubles strategy relies on coordinated positioning and reaction to the opponents' movements. A dedicated annotation tool capable of capturing synchronised player trajectories and tactical formations could significantly enhance tracking models in tennis doubles.

\textbf{Action Recognition and Shot Classification:}
Action recognition models aim to classify tennis shots (e.g., forehand, backhand, volley) based on video data. Gao \& Ju (2024) developed a deep learning-based tennis action recognition model using an attention-enhanced GRU (Gated Recurrent Unit) network trained on the Three Dimensional Tennis Shots (THETIS) dataset. Their system achieved high accuracy in classifying different stroke types, highlighting the potential of machine learning in automated shot classification.

However, existing action recognition datasets primarily focus on singles tennis. Doubles-specific action recognition presents additional complexities, such as interactions between teammates, synchronised shot sequences, and coordinated net play. A structured doubles annotation framework could facilitate the development of more sophisticated action recognition models capable of analysing team-based shot patterns and tactical sequences.

\textbf{Strategic Analysis:}
Tennis strategy modeling has advanced with machine learning and probabilistic approaches. Liu et al. (2024) introduced a Markov Decision Process (MDP)-based framework to analyse tennis doubles strategy, incorporating transition probabilities between tactical formations. This approach enables match outcome prediction based on historical doubles match data. Similarly, Martínez-Gallego et al. (2021) analysed volley positions in professional doubles matches using Hawk-Eye (a type of ball-tracking system) ball-tracking data, identifying distinct tactical patterns in men's and women's doubles.

These studies highlight the potential for data-driven strategic analysis in doubles tennis. However, access to high-resolution tracking data remains limited outside professional tournaments. A custom annotation tool could help generate structured datasets for analysing doubles tactics at various skill levels, enabling pattern recognition, formation analysis, and AI-driven coaching recommendations.

\section{Potential Tools to be integrated into the Labelling Tool}

\textbf{Player \& Object Tracking}
\textbf{Single Tennis Video Annotation:} Jiang, Izadi, Liu \& Dong (2020) developed a comprehensive automated annotation system for tennis broadcast videos that demonstrates significant potential for adaptation to doubles tennis analysis. Their system integrates multiple deep learning components for automated match analysis, achieving remarkable accuracy in player tracking (99.9\% precision, 99.5\% recall) and shot classification (92.9\% accuracy for five distinct shot types). The framework combines audio-visual shot detection, court mapping through homography transformation, advanced player tracking using Detectron2 with Simple Online and Realtime Tracking (SORT) algorithm, and sophisticated ball tracking through TrackNet, particularly effective for fast-moving or blurry ball detection. While their work focuses on singles tennis, their methodological approach to player movement analysis using skeletal key points (wrists, elbows) and their solution for handling player occlusion provides valuable insights for doubles tennis analysis. Their system's ability to convert video data into detailed shot-by-shot analytics demonstrates the feasibility of automated tennis analysis, though adaptation would be needed to address the unique challenges of doubles play, such as tracking four players simultaneously and analysing partner-to-partner interactions.

\textbf{Zero-shot Phrase Grounding:} Florence-2, Microsoft's unified vision foundation model, was introduced by Xiao et al. (2023), a pioneering unified vision foundation model that demonstrates exceptional capabilities in phrase grounding and object detection, which happens to be key components for automated tennis analysis. The model achieves state-of-the-art performance in visual grounding tasks (84.4\% R@1 on Flickr30k) and shows strong object detection capabilities (37.5 mAP on COCO) in zero-shot settings, making it particularly promising for tennis applications where annotated data is scarce. Their novel approach unifies spatial hierarchy (from image-level to pixel-level understanding) and semantic granularity (from coarse to fine-grained descriptions) through a unified sequence-to-sequence architecture. The fine-tuned base or large models can potentially be integrated to identify players through their provided descriptions.

Similar to Florence-2, Liu et al. (2023) introduced Grounding DINO, a novel open-set object detector that achieves state-of-the-art performance in visual grounding without requiring predefined object categories, making it particularly valuable for tennis applications. The model uniquely combines DINO architecture with grounded pre-training, enabling detection of arbitrary objects through text prompts or referring expressions. The end result is the same, which produces bounding boxes based on a text caption. The players can then be tracked throughout the sequence and rallies based on the given visual description.

\textbf{Pose Estimation Models}
\textbf{2D Pose Estimation:} In a paper published by Maji, Nagori, Mathew, \& Poddar in 2022, YOLO-Pose was introduced as a groundbreaking approach to multi-person pose estimation that is particularly relevant for tennis analysis. The system achieves state-of-the-art results (90.2\% AP50) while addressing several key challenges that are crucial for doubles tennis analysis. Unlike traditional two-stage approaches, YOLO-Pose can detect and track multiple players simultaneously in a single forward pass, making it computationally efficient for processing four players in doubles matches. Their novel heatmap-free approach uses standard object detection post-processing instead of complex post-processing steps, which is particularly advantageous for real-time tennis analysis. A significant innovation is their Object Keypoint Similarity (OKS) loss function, which directly optimises the evaluation metric instead of using surrogate losses, leading to more accurate pose estimation. However, their work, while promising, focuses on general pose estimation and would need adaptation for tennis-specific movements and the unique challenges of tracking four players simultaneously on a tennis court.

\textbf{3D Pose Estimation:} Mehraban, Adeli, \& Taat (2023) proposed MotionAGFormer, a novel architecture combining transformer and graph convolutional networks for 3D human pose estimation, which offers promising applications for tennis analysis. Their approach addresses several critical challenges in pose estimation that are particularly relevant to doubles tennis: handling multiple players simultaneously, capturing both global and local motion patterns, and maintaining accuracy during rapid movements. The model's dual-stream architecture, utilising both transformers for global relationships and GCNFormers (Graph Convolutional Networks + Transformer) for local joint dependencies, is especially useful and valuable for analysing the complex interactions and movements in doubles tennis. The system achieves state-of-the-art results (38.4mm MPJPE on Human3.6M dataset) while using significantly fewer parameters than previous approaches. Their approach to temporal modelling using a Graph Convolutional Network (GCN) that establishes connections between similar joint positions across frames could be particularly valuable for analysing tennis-specific movements like serves and volleys. However, while their work shows excellent use for pose estimation, it would need adaptation to handle the specific challenges of tennis court environments and the simultaneous tracking of four players in doubles matches. As of the current implementation, only one player can be tracked in each video.

\textbf{Pose Classification \& Shot Analysis}
\textbf{Action Recognition with GCNs :} Spatial-Temporal Graph Convolutional Networks (ST-GCN) was introduced in 2018 by Yan, Xiong \& Lin, which presents a compelling approach for analysing human movements through skeleton data, particularly relevant for tennis shot classification. Their key innovation is treating human joints as a graph structure where connections are based on natural body linkages and temporal relationships between frames. This approach is especially valuable for tennis analysis as it can capture both the spatial relationships between body parts (crucial for understanding stroke mechanics) and their temporal evolution during shots. The model achieves strong results on complex action recognition tasks (81.5\% accuracy on NTU-RGB+D dataset) without requiring manual part assignment or predefined rules, making it likely adaptable to various tennis strokes. More importantly, the premise of analysing frames within the videos can be translated for our use cases.

In addition, recent advancements in skeletal-based action recognition have shown promising developments for sports analysis, with SkateFormer (Do \& Kim, 2024) presenting a novel approach particularly relevant for tennis motion analysis. The paper introduces a Skeletal-Temporal Transformer that efficiently captures both spatial relationships between body joints and temporal patterns in movement sequences, which can also be applicable to tennis doubles' use cases. Their partition-specific attention strategy divides movements into four key categories – local and global motions combined with physically neighbouring and distant joint relationships to successfully recognise action patterns.

\textbf{2D CNNs:} While Graph Convolutional Networks (GCNs) have been effective in modeling the spatial-temporal dynamics of tennis shots, 2D Convolutional Neural Networks (CNNs) offer a simpler yet powerful alternative for shot classification tasks. By processing individual frames or sequences of frames, 2D CNNs can learn spatial features pertinent to different tennis strokes.

In the context of tennis shot analysis, 2D CNNs can be employed to extract spatial features from each frame of a video sequence. These features are then fed into temporal models, such as Recurrent Neural Networks (RNNs) or Temporal Convolutional Networks (TCNs), to capture the temporal dynamics of the shots. This approach leverages the strength of CNNs in spatial feature extraction and combines it with temporal modelling techniques to achieve effective shot classification.

For instance, Liu et al. (2025) introduced the Fast, Frequent, Fine-grained (F3Set) dataset, which includes tennis videos annotated with fine-grained shot information. They proposed a method that utilises 2D CNNs for spatial feature extraction, followed by temporal modelling to analyse fast, frequent, and fine-grained events in tennis matches. This approach demonstrated the potential of 2D CNNs in handling complex shot classification tasks in tennis.

\chapter{Data Sources}
For the purpose of this project, we will be using data from two different sources. The first one is public professional matches (highlights), downloaded from Youtube video, as well as close-sourced semi-professional National Collegiate Athletic Association (NCAA) videos obtained directly from the Louisiana State University (LSU) team. Videos for professional matches are higher definition than NCAA ones and are usually shorter in duration.

We proceeded to label the following videos to be used in this project to train, validate or test our models.

\begin{table}[h]
\centering
\renewcommand{\arraystretch}{1.2}
\begin{tabular}{|c|p{7cm}|c|c|}
\hline
\textbf{Source} & \textbf{Title} & \textbf{Dataset} & \textbf{\# of Events} \\
\hline
\multirow{4}{*}{Professional} & Granollers\_Zeballos vs Arevalo\_Rojer \_ Toronto 2023 Doubles Semi-Finals & Train/Val & 101 \\
\cline{2-4}
& Nick Kyrgios\_Thanasi Kokkinakis vs Jack Sock\_John Isner \_ Indian Wells 2022 Doubles Highlights & Train/Val & 110 \\
\cline{2-4}
& Rajeev Ram\_Joe Salisbury vs Tim Puetz\_Michael Venus \_ Cincinnati 2022 Doubles Final & Train/Val & 141 \\
\cline{2-4}
& Salisbury\_Ram vs Krawietz\_Puetz \_ Toronto 2023 Doubles Semi-Finals & Test & 88 \\
\hline
\multirow{4}{*}{NCAA} & an7MXASRyI0 & Train/Val & 251 \\
\cline{2-4}
& eGFJAG-2jM8 & Train/Val & 247 \\
\cline{2-4}
& EMBw\_kXc574 & Train/Val & 243 \\
\cline{2-4}
& VUPKfQgXy8g & Test & 177 \\
\hline
\end{tabular}
\caption{Dataset Information and Distribution}
\label{tab:dataset}
\end{table}

For standardisation, I will be using a 70-30 split for train-validation set across all the events. However, to ensure that for each experimentation, there is no data leakage and severe overfitting, I will be holding out two unseen videos: Salisbury\_Ram vs Krawietz\_Puetz  \_ Toronto 2023 Doubles Semi-Finals for Professional and VUPKfQgXy8g for NCAA.

\chapter{Implementation}
\begin{figure}[H]
    \centering
    \includegraphics[width=1\textwidth]{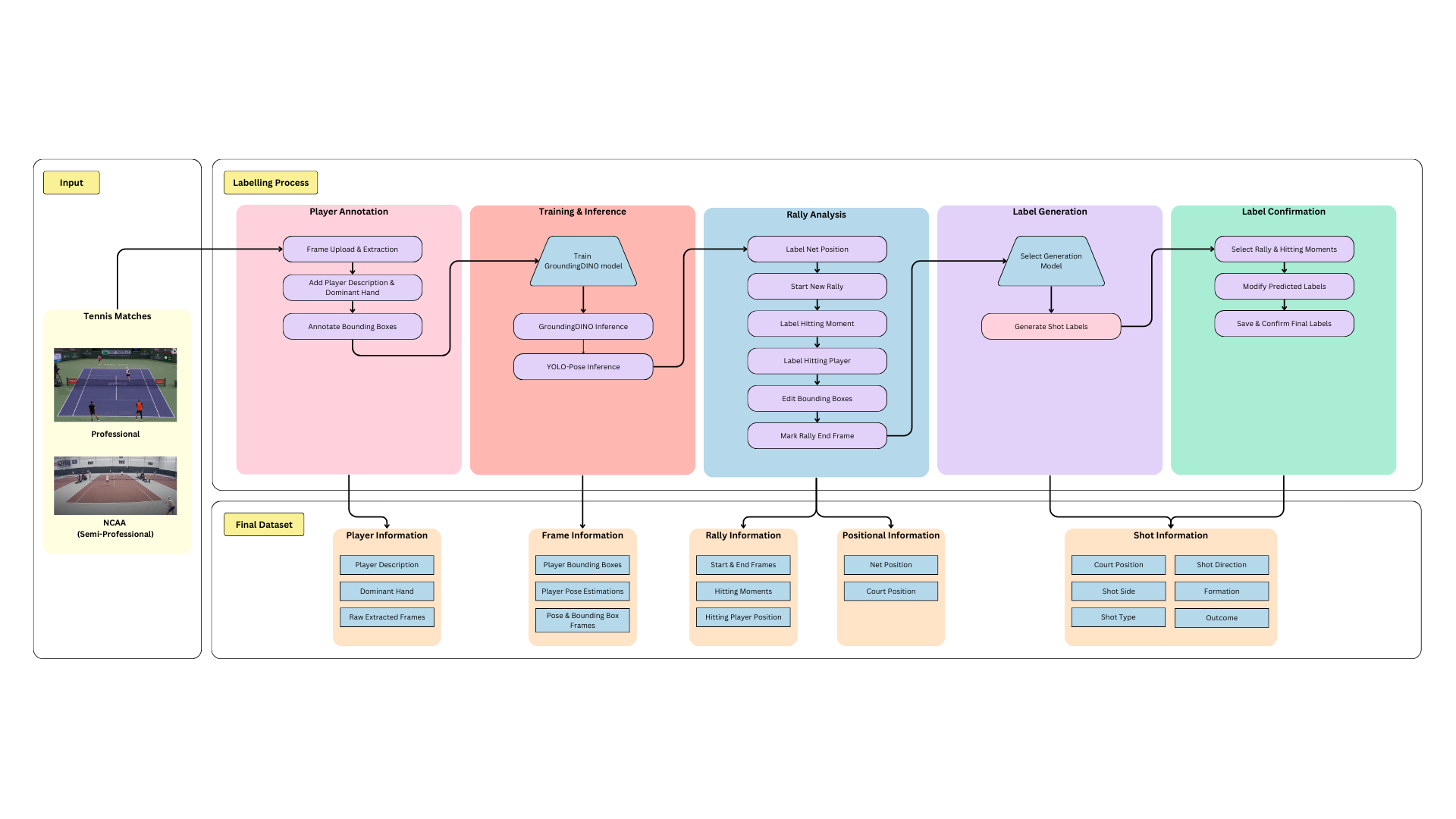}
    \caption{Tennis Double Annotation Tool Framework}
    \label{fig:annotation_framework}
\end{figure}

The above framework summarises the implementation of my final tennis annotation tool to produce structured, standardised and rich information for tennis doubles. This framework will incorporate the use of pre-trained machine learning models to automate and streamline some of the processes, such as Training \& Inference or Label Generation, greatly increasing efficiency of the labelling process.

\section{Standardised Video Annotation Framework}
The establishment of a standardised video annotation framework is crucial for systematically capturing and analysing data in doubles tennis. High-quality, structured data serves as the foundation for accurate performance analysis, tactical insights, and the development of automated systems. Collecting such data, however, presents significant challenges due to the dynamic nature of tennis doubles, including rapid player movements, complex positional shifts, and nuanced tactical formations. Without standardisation, annotations risk being inconsistent and incomplete, undermining their analytical utility and the reproducibility of findings. Consequently, a unified annotation framework is essential to ensure consistency, reliability, usefulness and comparability across datasets.

The proposed framework systematically organises data into four core categories: player information, positional information, frame information and rally information.

\textbf{Player Information:}
\begin{itemize}
\item Player description (Player 1, Player 2, Player 3, Player 4)
\item Dominant hand (left/right-handedness)
\end{itemize}

\textbf{Positional Information:}
\begin{itemize}
\item Net position
\item Player court position (Far Deuce, Far Advantage, Near Deuce, Near Advantage)
\end{itemize}

\textbf{Frame Information:}
\begin{itemize}
\item Player bounding boxes
\item Player pose estimations
\end{itemize}

\textbf{Rally Information:}
\begin{itemize}
\item Rally start and end frames
\item Number of hitting moments within each rally
\item Position of hitting player
\item Detailed hitting moment frame
\end{itemize}

\textbf{Shot Information} (for each hitting moment):
\begin{itemize}
\item Court Position: Far Deuce, Far Advantage, Near Deuce, Near Advantage
\end{itemize}

\begin{figure}[H]
    \centering
    \includegraphics[width=1\textwidth]{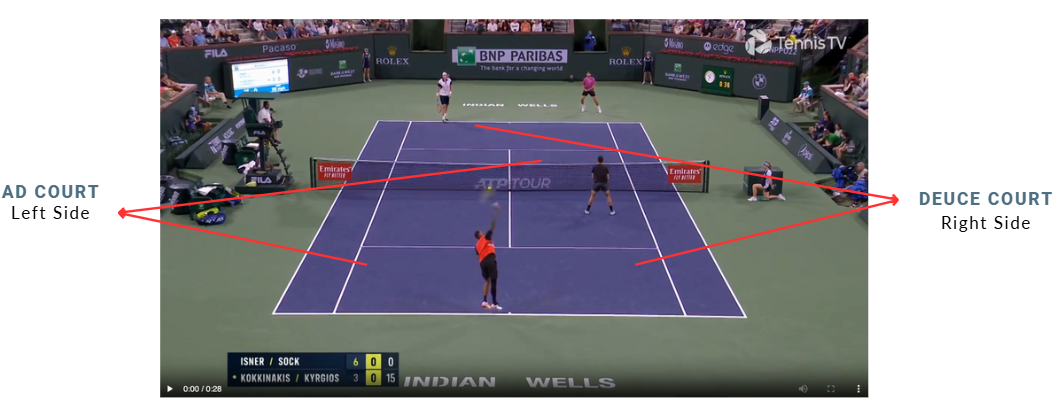}
    \caption{Court Position}
    \label{fig:court_position}
\end{figure}

For court position, we can use the net position, as well as player position marked during the rally analysis stage to determine whether he is standing in which quadrant of the court.

\begin{itemize}
\item Shot Side: Forehand, Backhand
\end{itemize}

\begin{figure}[H]
    \centering
    \includegraphics[width=1\textwidth]{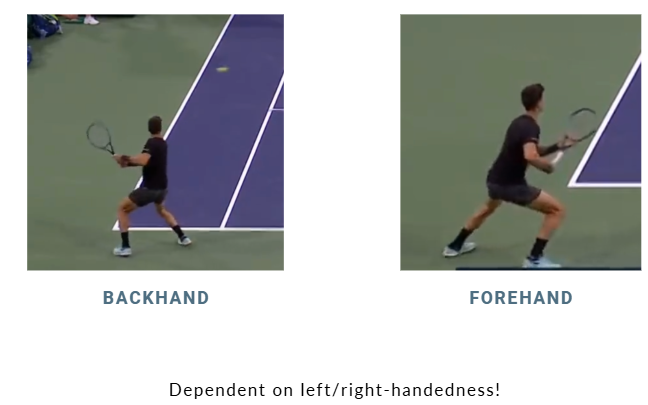}
    \caption{Shot Side}
    \label{fig:shot_side}
\end{figure}

\begin{itemize}
\item Shot Type: Serve, Second-Serve, Return, Volley, Lob, Smash, Swing
\end{itemize}

\begin{figure}[H]
    \centering
    \includegraphics[width=1\textwidth]{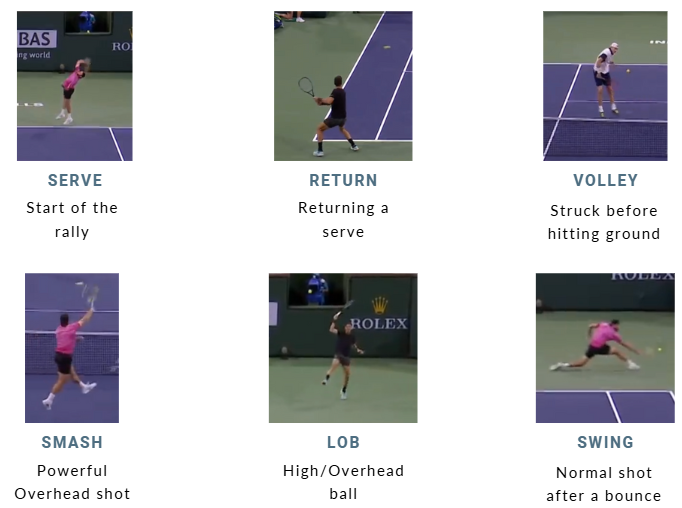}
    \caption{Shot Types}
    \label{fig:shot_types}
\end{figure}

\begin{itemize}
\item Shot Direction: 
   \begin{itemize}
   \item Serves: T, Body (B), Wide (W)
   \item Non-serves: Cross Court (CC), Down Line (DL), Inside-in (II), Inside-out (IO)
   \end{itemize}
\end{itemize}

\begin{figure}[H]
    \centering
    \includegraphics[width=1\textwidth]{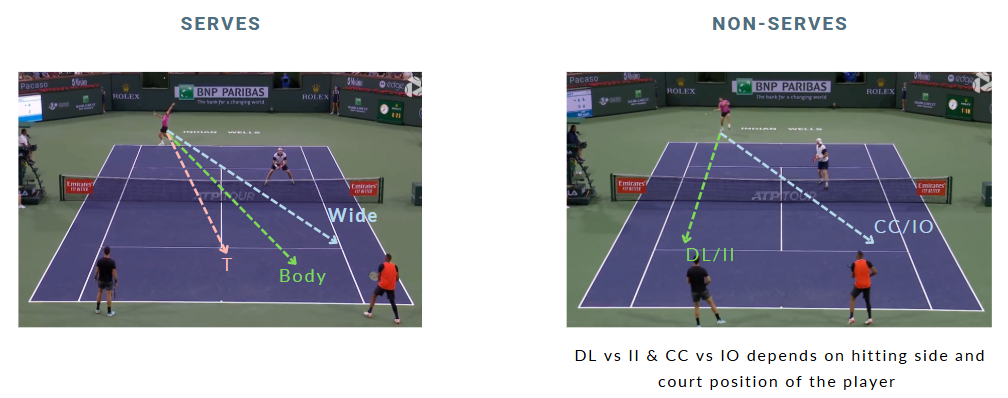}
    \caption{Shot Directions}
    \label{fig:shot_directions}
\end{figure}

Determination of DL vs II and CC vs IO is discussed in the above Section 4.1.

\begin{itemize}
\item \textbf{Formation:} Conventional, I-Formation, Australian, Non-serve
\end{itemize}

\begin{figure}[H]
    \centering
    \includegraphics[width=1\textwidth]{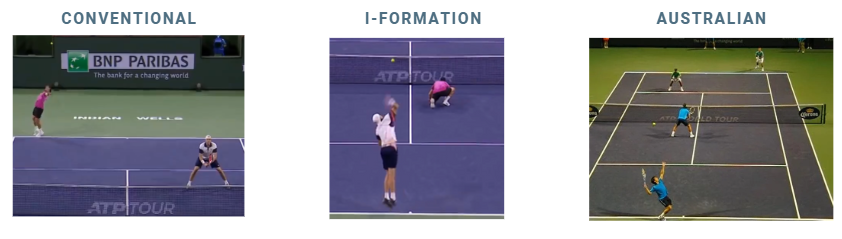}
    \caption{Formations}
    \label{fig:formations}
\end{figure}

\begin{itemize}
\item \textbf{Outcome:} In, Win, Error
\end{itemize}

Outcome is basically whether the ball is still in play (In), won by the hitting player (Win) or lost by the hitting player through unforced/forced errors (Error).

Additionally, specific labelling rules have been established to standardise annotation further:

\textbf{Shot-Direction-Formation Combinations:}
\begin{itemize}
\item Serves: Always annotated with T/B/W in conjunction with Conventional/I-Formation/Australian.
\item Non-serves: Annotated with CC/DL/II/IO combined with Non-serve formation.
\end{itemize}

\textbf{Court-Side-Direction Combinations:}
\begin{itemize}
\item Right-handed players:
   \begin{itemize}
   \item Advantage side: Backhand (CC/DL), Forehand (II/IO)
   \item Deuce side: Forehand (CC/DL), Backhand (II/IO)
   \end{itemize}
\item Left-handed players:
   \begin{itemize}
   \item Advantage side: Forehand (CC/DL), Backhand (II/IO)
   \item Deuce side: Backhand (CC/DL), Forehand (II/IO)
   \end{itemize}
\end{itemize}

Much of the information, such as the player's dominant hand, commonly not collected or noted in datasets, can be quite crucial for tennis double analytics. This standardisation will streamline the annotation process, enhance data quality, and lay a robust foundation for developing automated analytical models, aligning directly with the project's broader objectives of integrating automation into the analytical workflow.

\section{Tennis Annotation Tool}
\textbf{Source Code:} \href{https://github.com/alvinjiang1/tennis-annotation-tool}{\texttt{Tennis Annotation Tool}}

\textbf{Annotation Tool Guide:} \href{https://youtu.be/nXhNrud8wdU}{\texttt{Demo}}

The tennis annotation tool, user flow, requirements and implementation is developed jointly with Alvin.

\textbf{Functional Requirements}

\small
\begin{longtable}{|p{2.5cm}|p{3cm}|p{8cm}|}
\hline
\textbf{Page} & \textbf{Requirement} & \textbf{Description} \\ 
\hline
\endfirsthead

\multicolumn{3}{c}%
{\tablename\ \thetable\ -- \textit{Continued from previous page}} \\
\hline
\textbf{Page} & \textbf{Requirement} & \textbf{Description} \\ 
\hline
\endhead

\hline \multicolumn{3}{r}{\textit{Continued on next page}} \\ 
\endfoot

\hline
\endlastfoot

Player Annotation 
& Video Upload/Selection & Upload new videos or select previously uploaded videos for frame annotation. \\ \cline{2-3}
& Player Definition & Define player descriptions and handedness (left/right/unknown). \\ \cline{2-3}
& Frame Navigation & Navigate frames using buttons, slider, keyboard shortcuts. \\ \cline{2-3}
& Bounding Box Annotation & Annotate bounding boxes around players with category selection. \\ \cline{2-3}
& Annotation Management & View, edit, delete existing annotations. \\ \cline{2-3}
& Save Annotations & Save annotations in COCO format. \\ \cline{2-3}
& Frame Status Tracking & Indicate annotated frames visually. \\ \hline

Training \& Inference 
& Video Selection & Select annotated videos for training/inference. \\ \cline{2-3}
& Readiness Verification & Verify annotation completeness and frame extraction. \\ \cline{2-3}
& Training Execution & Run and monitor GroundingDINO training. \\ \cline{2-3}
& Training Status & Real-time training status updates. \\ \cline{2-3}
& Inference Execution & Run player detection inference on video frames. \\ \cline{2-3}
& Inference Status & Track inference progress and completion. \\ \cline{2-3}
& Model Management & Reset training environment for retraining. \\ \hline

Rally Analysis
& Video Loading & Load videos with inference results. \\ \cline{2-3}
& Net Position Marking & Set net position for court orientation. \\ \cline{2-3}
& Rally Creation & Specify start frames of rallies. \\ \cline{2-3}
& End Frame Marking & Define rally end frames and ball positions. \\ \cline{2-3}
& Hitting Moment Annotation & Mark player hitting frames and identify hitter. \\ \cline{2-3}
& Bounding Box Editing & Correct bounding boxes and re-run inference. \\ \cline{2-3}
& Rally Management & Manage rally data and hitting moments. \\ \cline{2-3}
& Frame Navigation & Navigate frames using UI controls. \\ \hline

Label Generation
& Video Selection & Select analysed videos for label generation. \\ \cline{2-3}
& Model Selection & Select prediction models (Random, CNN, Gemini). \\ \cline{2-3}
& Label Generation & Generate shot labels from rally data. \\ \cline{2-3}
& Label Visualization & View shot labels per rally. \\ \cline{2-3}
& Player Information Display & Show player descriptions and handedness. \\ \cline{2-3}
& Label Format Configuration & Display label component details. \\ \hline

Label Review 
& Video Selection & Load videos with shot labels. \\ \cline{2-3}
& Frame Navigation & Navigate labelled frames. \\ \cline{2-3}
& Label Visualization & Detailed shot label display. \\ \cline{2-3}
& Label Editing & Edit shot properties. \\ \cline{2-3}
& Validation Enforcement & Ensure tennis rules compliance. \\ \cline{2-3}
& Frame Preview & Preview shot execution frames. \\ \cline{2-3}
& Label Source Tracking & Track generated vs. confirmed labels. \\ \cline{2-3}
& Label Confirmation & Confirm and save shot labels. \\ \hline

Miscellaneous
& Navigation System & Sidebar navigation between main pages. \\ \cline{2-3}
& Theme Support & Switch visual themes. \\ \cline{2-3}
& User Guidance & Detailed help sections. \\ \cline{2-3}
& Keyboard Shortcuts & Support keyboard navigation/actions. \\ \cline{2-3}
& Error Handling & Informative and graceful error handling. \\ 
\hline
\caption{Functional Requirements}
\label{tab:functional_requirements}
\end{longtable}

\textbf{Non-Functional Requirements}
\small
\begin{longtable}{|p{2.5cm}|p{3cm}|p{8cm}|}
\hline
\textbf{Category} & \textbf{Requirement} & \textbf{Description} \\ 
\hline
\endfirsthead

\multicolumn{3}{c}%
{\tablename\ \thetable\ -- \textit{Continued from previous page}} \\
\hline
\textbf{Category} & \textbf{Requirement} & \textbf{Description} \\ 
\hline
\endhead

\hline \multicolumn{3}{r}{\textit{Continued on next page}} \\ 
\endfoot

\hline
\endlastfoot

Performance
& Response Time & Respond within 200ms to maintain smooth interactions (except training/inference). \\ \cline{2-3}
& Frame Loading & Load frames within 0.5 seconds when navigating. \\ \cline{2-3}
& Processing Time & Inference completed within 5 mins per 100 frames. \\ \cline{2-3}
& Resource Utilization & Efficient system resource usage (CPU/GPU/memory). \\ \hline

Usability
& Intuitive Interface & Interface usable by individuals with basic computer skills. \\ \cline{2-3}
& Error Messages & Clear, specific, and actionable errors. \\ \cline{2-3}
& Consistency & Consistent UI/navigation/terminology. \\ \hline

Reliability
& Data Preservation & Preserve data during unexpected shutdowns. \\ \cline{2-3}
& Error Recovery & Graceful error recovery without data loss. \\ \cline{2-3}
& Crash Rate & Fewer than 1 crash per 8 continuous hours. \\ \hline

Security
& Data Protection & Secure data storage with appropriate access controls (currently local deployment). \\ \hline

Maintainability
& Code Quality & Follow consistent coding standards. \\ \cline{2-3}
& Documentation & Clear code comments and documentation. \\ \cline{2-3}
& Modularity & Modular design for easier maintenance and upgrades. \\ 
\hline
\caption{Non-Functional Requirements}
\label{tab:nonfunctional_requirements}
\end{longtable}
\clearpage

\subsection{Player Annotation Page}
The Player Annotation Page serves as the entry point for the annotation workflow, allowing users to upload new videos or select previously uploaded ones for frame extraction and player identification. Users can define player descriptions with associated handedness attributes and draw bounding boxes around players in each frame, establishing the foundation for subsequent model training.

\begin{figure}[H]
    \centering
    \includegraphics[width=1\textwidth]{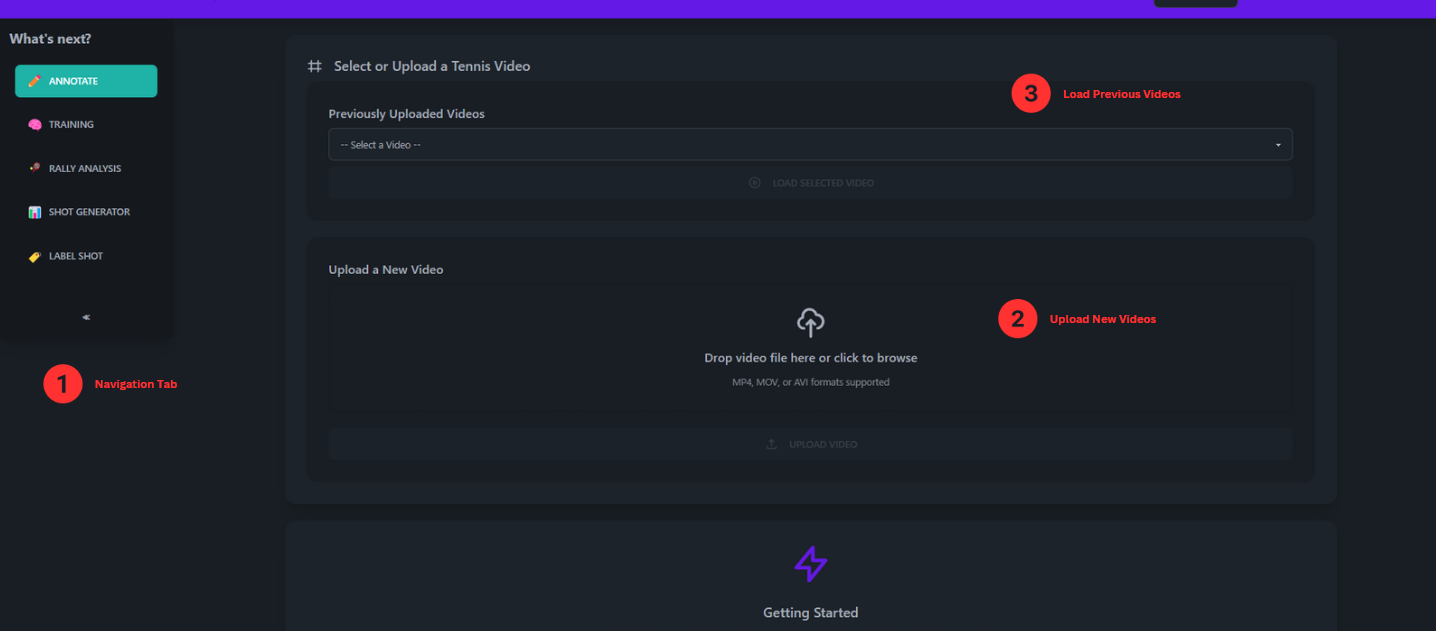}
    \includegraphics[width=1\textwidth]{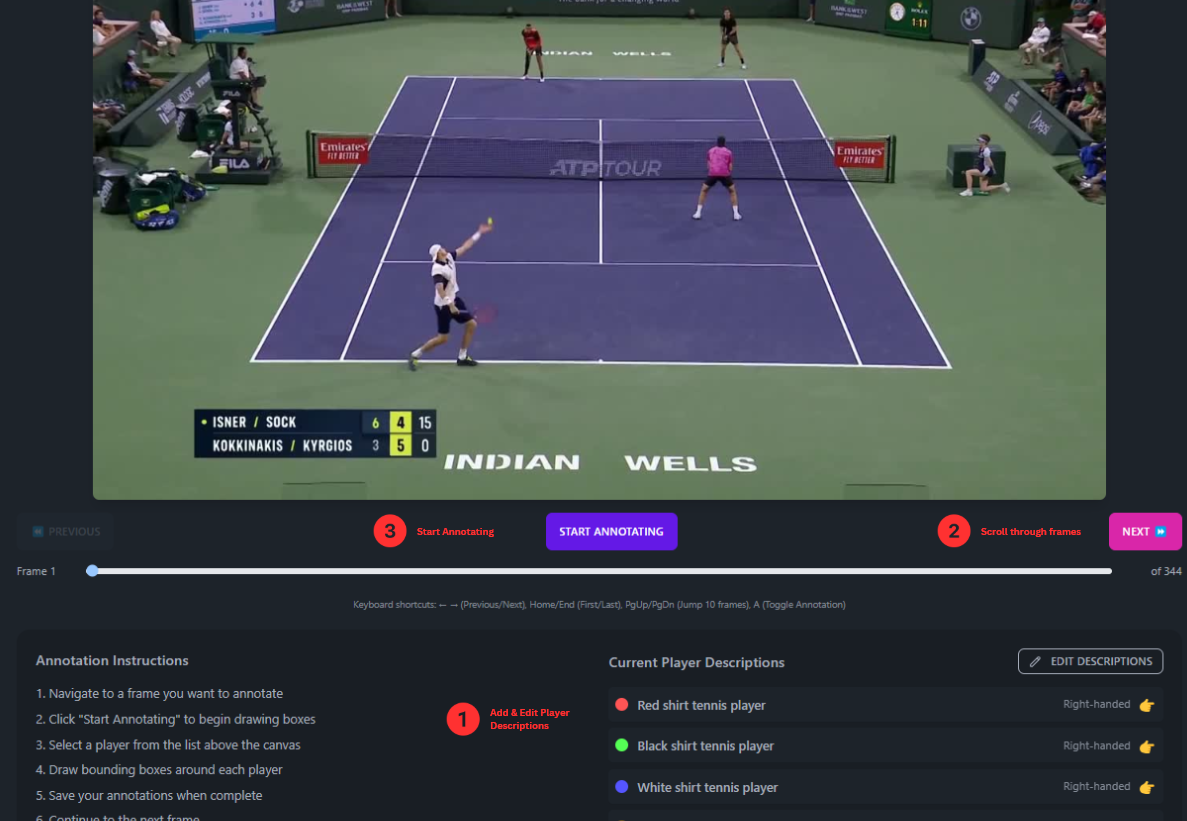}
    \caption{Player Annotation Page Interface}
    \label{fig:player_annotation}
\end{figure}

\begin{figure}[H]
    \centering
    \includegraphics[width=1\textwidth]{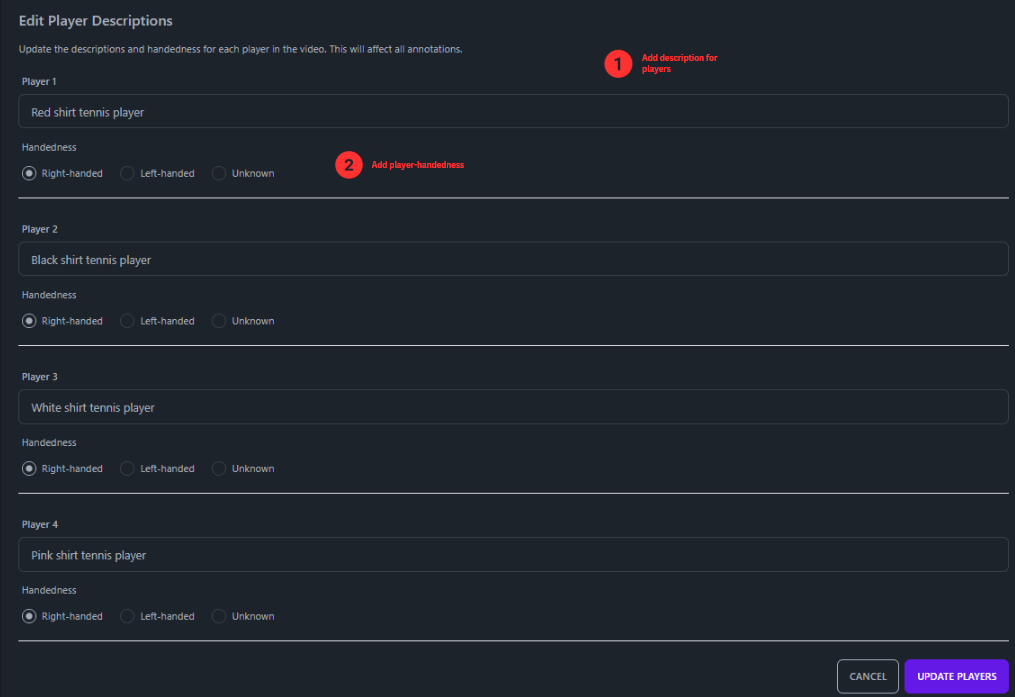}
    \includegraphics[width=1\textwidth]{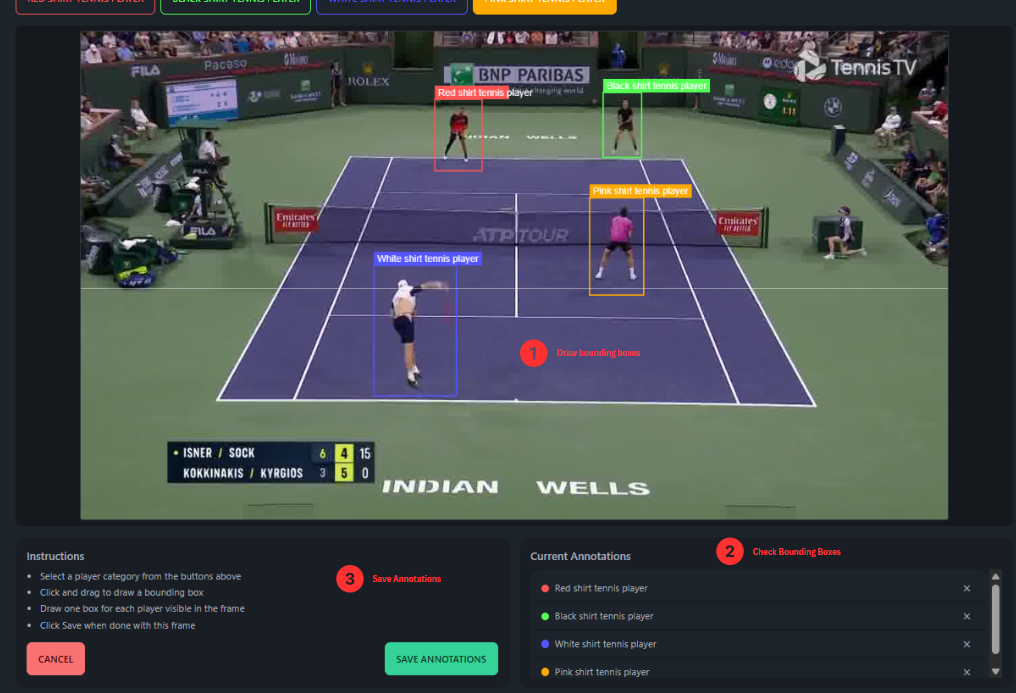}
    \caption{Player Bounding Box Interface}
    \label{fig:player_bounding_box}
\end{figure}
\clearpage

\subsection{Training \& Inference Page}
The Training \& Inference Page enables users to train the GroundingDINO model using previously annotated videos, with verification systems set in place to ensure videos are properly prepared with necessary annotations and extracted frames. Real-time status updates keep users informed of the training and inference progress, allowing them to monitor the model's development. Once training is complete, users can run inference (YOLO-Pose \& GroundingDINO) to detect players across all video frames, establishing the groundwork for rally analysis and shot labelling.

\begin{figure}[H]
    \centering
    \includegraphics[width=1\textwidth]{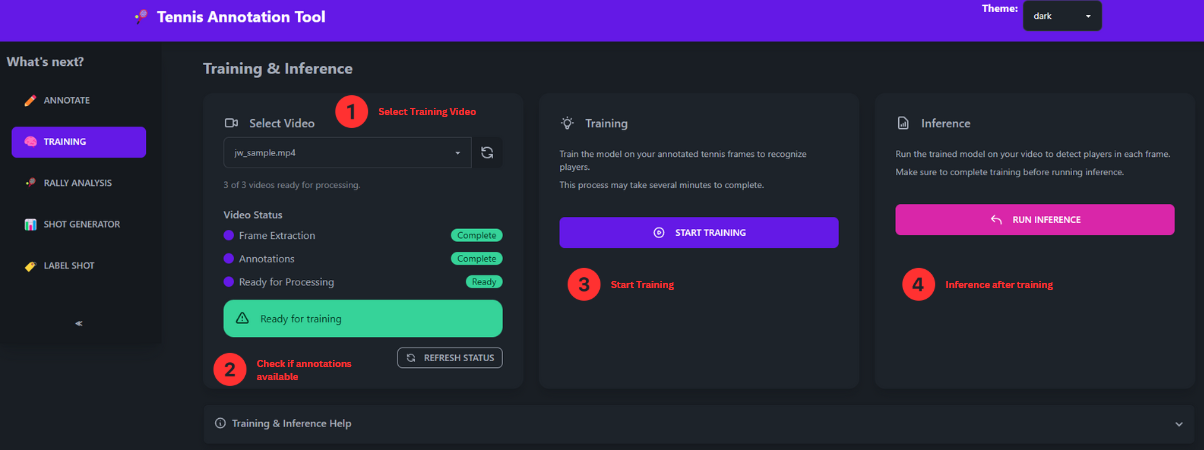}
    \caption{Training and Inference Page Interface}
    \label{fig:training_inference}
\end{figure}
\clearpage

\subsection{Rally Analysis Page}
The Rally Analysis Page allows users to identify and mark complete tennis rallies within processed videos, including setting the net position to establish court positions. Users can then create multiple rallies, annotate hitting moments with hitting player's position, and edit incorrect bounding boxes when necessary, with all changes triggering automated re-inference on affected frames.

\begin{figure}[H]
    \centering
    \includegraphics[width=1\textwidth]{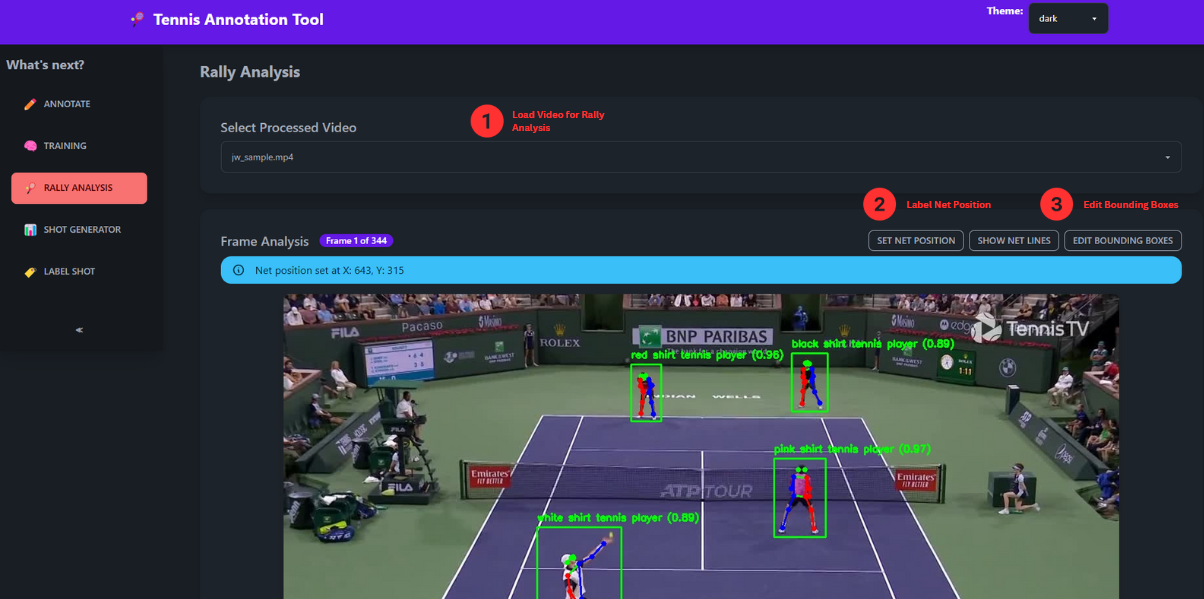}
    \includegraphics[width=1\textwidth]{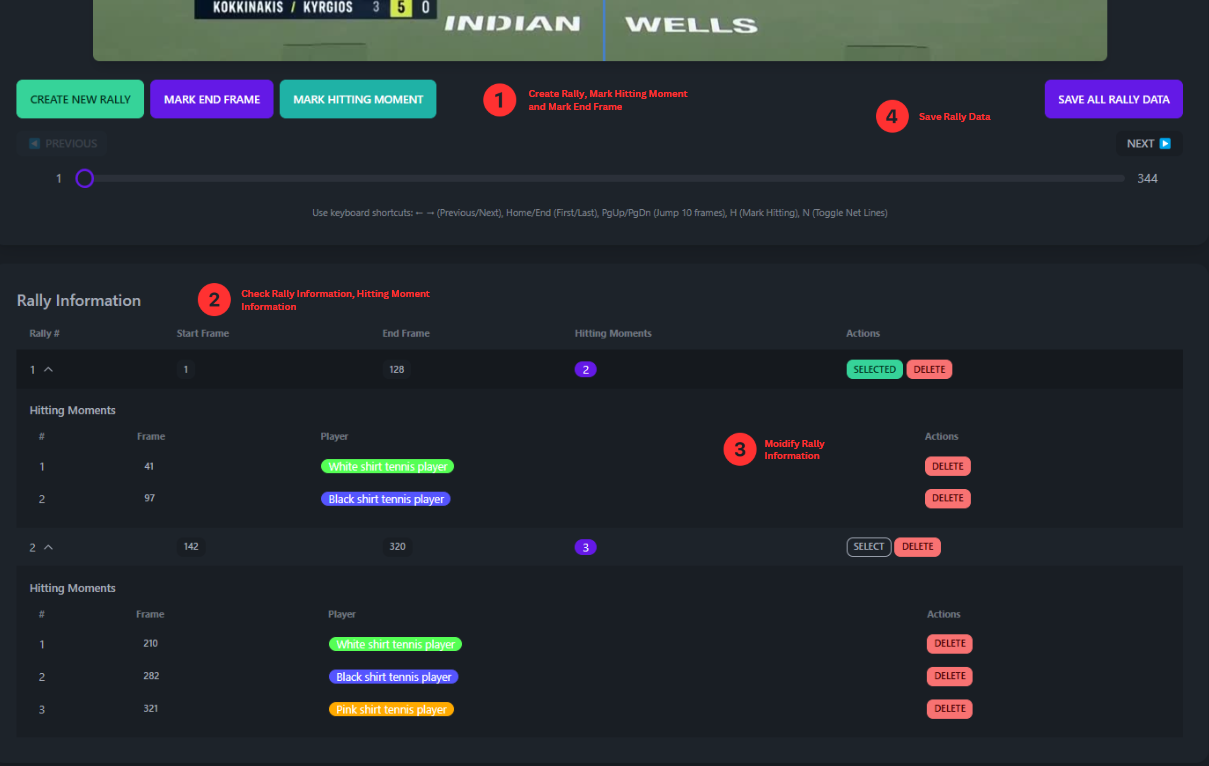}
    \caption{Rally Analysis Page Interface}
    \label{fig:rally_analysis}
\end{figure}
\clearpage

\subsection{Label Generation Page}
The Label Generation Page automates the process of creating detailed shot labels based on the rally analysis data, allowing users to select from different prediction models including Random, CNN, and Gemini. Generated labels incorporate vital information such as court position, shot type, technique, direction, formation, and outcome, providing a comprehensive classification of each tennis shot. The interface displays the generated labels in an organised format, allowing users to review the results.

\begin{figure}[H]
    \centering
    \includegraphics[width=1\textwidth]{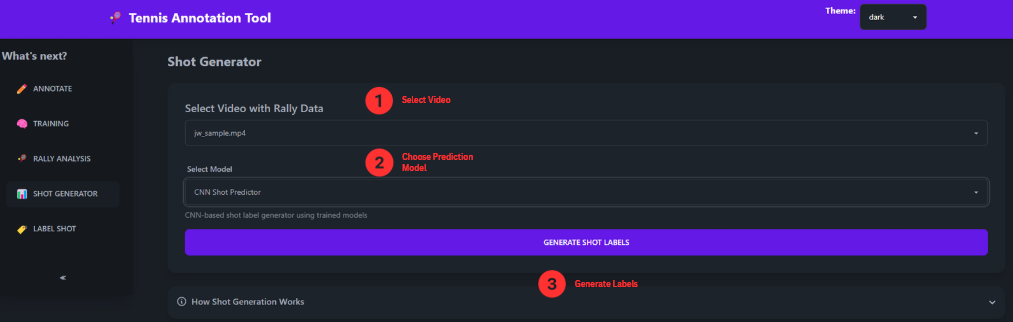}
    \includegraphics[width=1\textwidth]{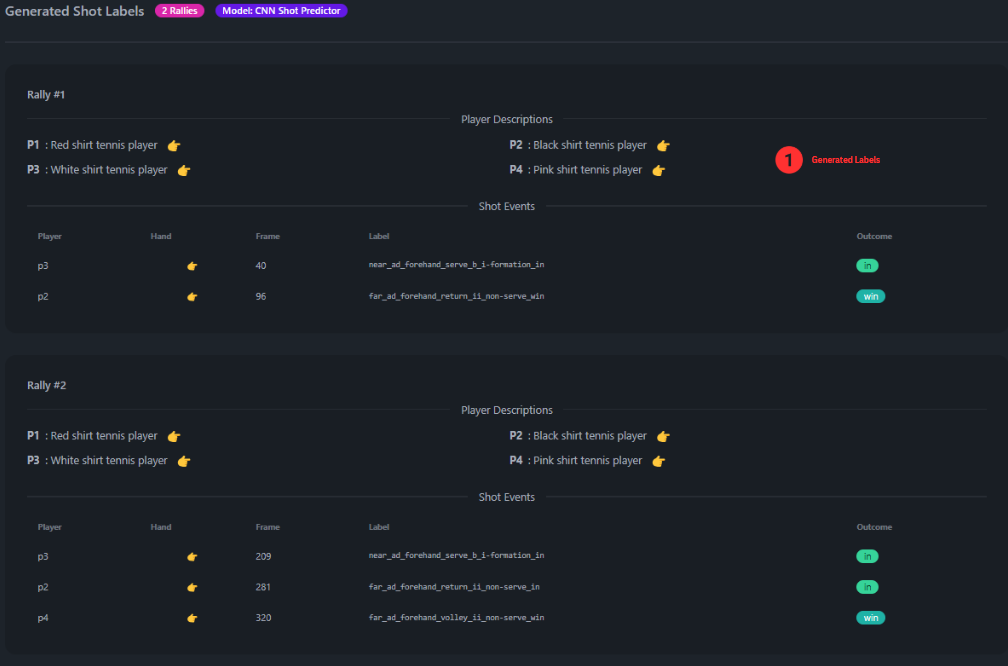}
    \caption{Label Generation Page Interface}
    \label{fig:label_generation}
\end{figure}
\clearpage

\subsection{Label Confirmation Page}
The Label Confirmation Page enables users to review, modify, and confirm automatically generated shot labels, with built-in validation rules based on tennis strategy and player handedness to ensure accuracy. Users can navigate through frames with a frame preview feature that allows scrolling ahead or back to observe ball trajectory. After confirmation, the final labels can be regarded as a reliable dataset.
\begin{figure}[H]
    \centering
    \includegraphics[width=1\textwidth]{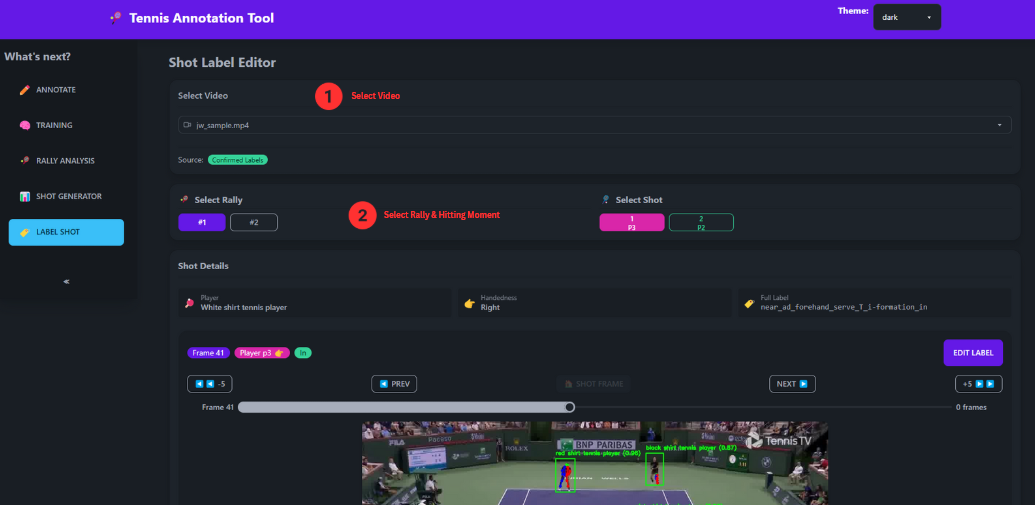}
    \includegraphics[width=1\textwidth]{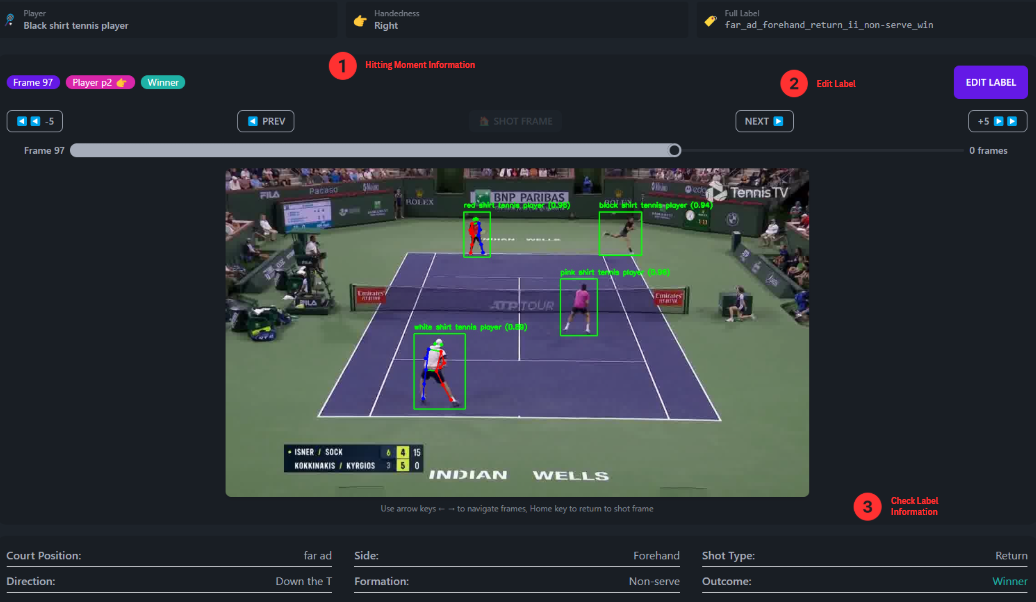}
    \caption{Label Confirmation Page Interface}
    \label{fig:label_confirmation}
\end{figure}
\clearpage

\subsection{Application Optimisations}
As part of optimising the tennis annotation tool, I have implemented several optimisations to supplement several non-functional requirements, enhancing performance, resource utilisation, and user experience across both frontend and backend components:

\textbf{Frontend Optimisations}
\begin{enumerate}
    \item \textbf{Responsive UI Architecture:} The application employs a responsive design framework using Tailwind CSS and DaisyUI components, automatically adapting to different screen sizes without sacrificing functionality. Layout components dynamically adjust based on available space, maintaining usability across devices.
    
    \item \textbf{State Management Efficiency:} Custom hooks like useToast, useVideos, and useFrames encapsulate complex state logic, reducing unnecessary re-renders and improving component efficiency. This ensures that only components affected by state changes are updated, conserving resources.
    
    \item \textbf{Lazy Loading \& Caching:} Frame navigation implements optimised image loading with caching to ensure fresh data when needed for performance. The system prioritizes loading only the current frame at full resolution, with placeholder mechanisms to maintain UI responsiveness.
\end{enumerate}

\textbf{Backend Optimisations}
\begin{enumerate}
    \item \textbf{Model Hot-Loading:} The CNN model implementation utilises a hot-loading system that keeps trained models in memory between inference requests, eliminating initialisation overhead and significantly reducing processing time for shot label prediction. The tradeoff results in a slightly longer startup time and higher GPU usage for the backend, which in our scenario, is better than re-loading all models on inference. The following result is tested with the inference of 54 rallies:
    
    \begin{table}[H]
        \centering
        \begin{tabular}{|l|c|c|c|}
        \hline
        \textbf{Method} & \textbf{Startup Time} & \textbf{Inference Time} & \textbf{GPU Usage (MAX)} \\
        \hline
        Cold Start & 5s & 12 min 35s & 0.8 GB \\
        \hline
        Hot Loading & 20s & 54s & 3.2 GB \\
        \hline
        \end{tabular}
        \caption{Cold-start vs Hot-Loading of CNN models}
        \label{tab:model_loading}
    \end{table}
    
    \item \textbf{Parallel Processing for Inference:} GroundingDINO frame extraction leverages multi-threading to process frames in parallel, achieving near-linear speed-up with multiple cores available in the system, greatly reducing processing time especially for longer videos.
    
    \item \textbf{COCO Format Integration:} Annotations are stored in the standardised COCO format to reduce conversion overhead during the annotation, training and inference processes.
    
    \item \textbf{Asynchronous Processing Pipelines:} Resource-intensive operations like GroundingDINO inference and pose estimation are processed asynchronously in the backend, allowing users to continue working with the application while these operations run in the background.
\end{enumerate}

\section{Automated Tennis Analyses}
As discussed in the framework shown in Figure \ref{fig:annotation_framework}, manually annotating some of the tennis doubles videos can be extremely time-consuming and tedious. These scenarios include extracting the relevant bounding boxes and pose estimations of each player across all frames in the training \& inference stage, as well as manually generating all the labels (shot type, shot direction, formation, etc.) in the shot labelling stage. Therefore, as part of improving the tennis labelling workflow, we have developed methods incorporating machine learning models to minimise the time taken to extract the data.

\subsection{Bounding Boxes \& Pose Estimation}
To enable robust player tracking across tennis match rallies, we tested a two-stage detection and tracking pipeline combining YOLOv11 (Khanam, R., \& Hussain, M, 2024) and Deep SORT (Wojke, Bewley, \& Paulus, 2017). This approach allows us to maintain consistent player identities throughout rallies, even during occlusions or brief detection failures.

\begin{figure}[H]
   \centering
   \includegraphics[width=1\textwidth]{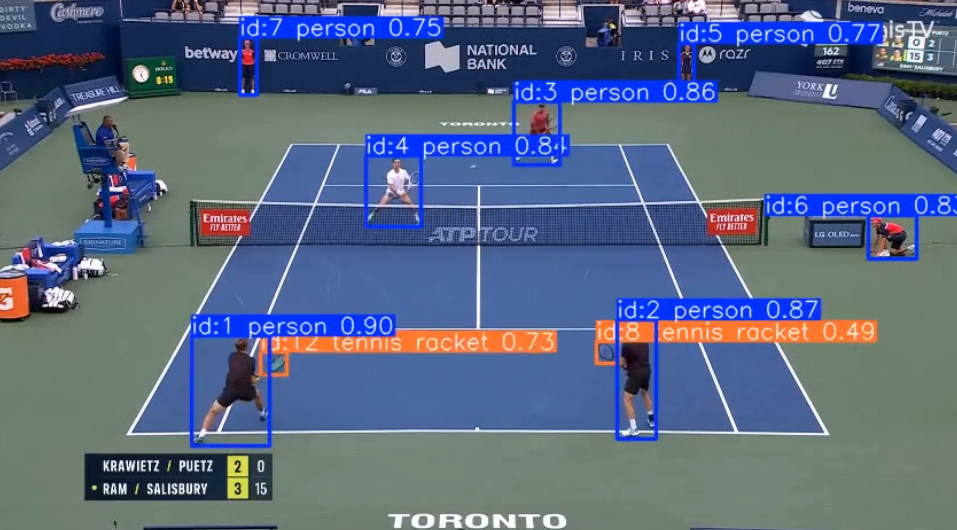}
   \caption{Successful Player Tracking with YOLOv11 \& DeepSORT algorithm}
   \label{fig:successful_tracking}
\end{figure}

While this pipeline effectively tracks players during unobstructed play, it faces challenges with occlusions, far-court players, and identity switching between rallies. The system's performance is particularly dependent on video quality and camera positioning, with tracking accuracy diminishing significantly for distant players. This is evident in Figure \ref{fig:failed_tracking} shown below, where we are unable to accurately obtain 2D poses from players that are not tracked by the YOLO models:

\begin{figure}[H]
   \centering
   \includegraphics[width=1\textwidth]{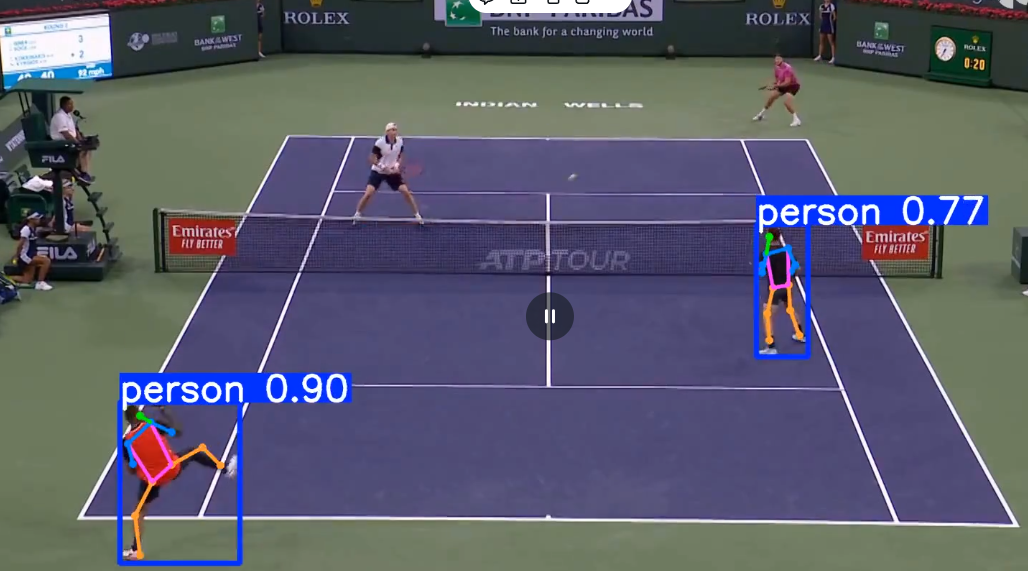}
   \caption{Failed far-court Player Tracking \& Pose Estimation via YOLOv11 \& DeepSORT algorithm}
   \label{fig:failed_tracking}
\end{figure}

To improve detection reliability, we implemented a two-stage pipeline combining Grounding DINO (Liu et al., 2023) and YOLO-Pose (Maji et al., 2022). Grounding DINO first performs zero-shot detection using natural language prompts (``tennis player'') to precisely localise players, effectively filtering out irrelevant detections. These localised regions are then processed by YOLO-Pose to extract 2D key-point coordinates for each player. This approach significantly improves pose estimation consistency by eliminating false detections through semantic filtering, providing higher-quality input regions for pose estimation. In addition, using a phrase grounding model such as GroundingDINO, we are able to differentiate between different players based on their description, greatly enriching the data collected.

\begin{figure}[H]
   \centering
   \includegraphics[width=1\textwidth]{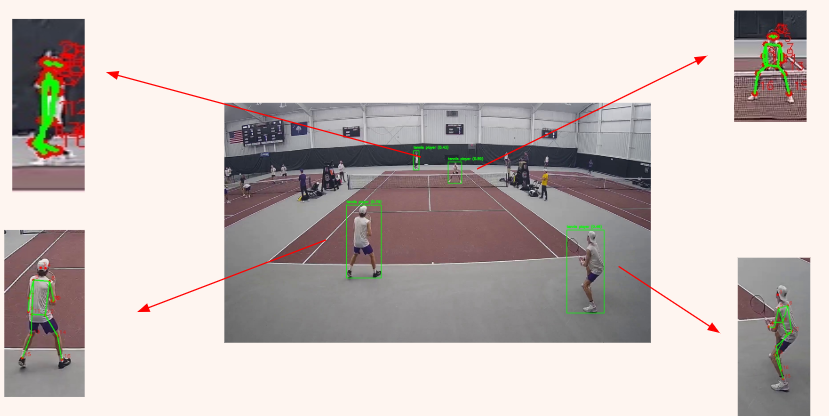}
   \caption{Successful Player Localization \& Pose Estimation via Grounding DINO \& YOLO-Pose}
   \label{fig:successful_localization}
\end{figure}

As shown in Figure \ref{fig:successful_localization}, with a fine-tuned Grounding DINO model, we are able to more accurately track far players and only track relevant tennis players. Despite some of the frames still tracking more than four players, we can perform several video-processing techniques, such as court cropping, to further enhance the improvements. This pipeline shows promising results if we continually finetune and improve the base Grounding DINO model.

\vspace{1cm}
\textbf{Bounding Boxes \& Pose Estimation Results}

This section presents a comparative analysis of various models in three key tasks: object tracking, phrase grounding, and pose estimation. Our evaluation metrics focus on player detection accuracy (percentage of successful detections per frame) and computational efficiency (processing time per rally).

P1, P2, P3, P4 refers to bottom left, bottom right, top left and top right players respectively. Player identification refers to whether the model can differentiate between P1, P2, P3 and P4 based on their visual descriptions.

In the first video, we are looking at a rally from a NCAA doubles match that spans 14 seconds and has a total of 952 frames. This rally is from a non-professional match that has slightly poorer video quality.

\begin{table}[H]
    \centering
    \small
    \begin{tabular}{|>{\centering\arraybackslash}p{2.5cm}|>{\centering\arraybackslash}p{2.5cm}|>{\centering\arraybackslash}p{1cm}|>{\centering\arraybackslash}p{1cm}|>{\centering\arraybackslash}p{1cm}|>{\centering\arraybackslash}p{1cm}|>{\centering\arraybackslash}p{1.7cm}|>{\centering\arraybackslash}p{1.5cm}|}
    \hline
    \textbf{Category} & \textbf{Model} & \textbf{P1 (\%)} & \textbf{P2 (\%)} & \textbf{P3 (\%)} & \textbf{P4 (\%)} & \textbf{Time taken (s)} & \textbf{Identify Player} \\
    \hline
    Object Tracking & YOLOv11 \& DeepSORT & 100 & 95.6 & 43.4 & 0.1 & 130 & No \\
    \hline
    Phrase Grounding & Florence-2 Large & 87.3 & 85.6 & 0.1 & 0.0 & 2150 & Yes \\
    \hline
    \multirow{2}{*}{Pose Estimation} & YOLO-Pose & 100 & 94.3 & 0.0 & 0.0 & 170 & No \\
    \cline{2-8}
    & Grounding DINO + YOLO-Pose & 100 & 100 & 69.8 & 64.3 & 1436 & Yes \\
    \hline
    \end{tabular}
    \caption{Performance Evaluation on NCAA Doubles Match (Rally \#1)}
    \label{tab:ncaa_performance}
\end{table}
\clearpage

In the second video, we are looking at a rally of Nick Kyrgios \& Thanasi Kokkinakis vs Jack Sock \& John Isner during Indian Wells 2022. This match spans 7 seconds and has a total of 451 frames, which is also of a better video quality. 

\begin{table}[H]
    \centering
    \small
    \begin{tabular}{|>{\centering\arraybackslash}p{2.5cm}|>{\centering\arraybackslash}p{2.5cm}|>{\centering\arraybackslash}p{1cm}|>{\centering\arraybackslash}p{1cm}|>{\centering\arraybackslash}p{1cm}|>{\centering\arraybackslash}p{1cm}|>{\centering\arraybackslash}p{1.7cm}|>{\centering\arraybackslash}p{1.5cm}|}
    \hline
    \textbf{Category} & \textbf{Model} & \textbf{P1 (\%)} & \textbf{P2 (\%)} & \textbf{P3 (\%)} & \textbf{P4 (\%)} & \textbf{Time taken (s)} & \textbf{Identify Player} \\
    \hline
    Object Tracking & YOLOv11 \& DeepSORT & 100 & 100 & 95.3 & 95.6 & 65 & No \\
    \hline
    Phrase Grounding & Florence-2 Large & 100 & 100 & 70.9 & 73.8 & 1022 & Yes \\
    \hline
    \multirow{2}{*}{Pose Estimation} & YOLO-Pose & 100 & 100 & 91.1 & 89.1 & 81 & No \\
    \cline{2-8}
    & Grounding DINO + YOLO-Pose & 100 & 100 & 89.4 & 84.1 & - & Yes \\
    \hline
    \end{tabular}
    \caption{Performance Evaluation on YouTube Highlights (Rally \#3)}
    \label{tab:youtube_performance}
\end{table}

The vision models show a significant variation in performance based on both the video quality and the task complexity. For object tracking, the YOLOv11 \& DeepSORT combination consistently achieves high accuracy for near-side players (P1, P2) across both lower-quality NCAA matches and professional-grade footage, showcasing the robustness of YOLO-based tracking methods to detect prominently positioned players even under suboptimal conditions. However, performance notably deteriorates when tracking far-side players (P3, P4) in the lower-quality NCAA video, highlighting the sensitivity of these algorithms to environmental factors such as lightning and video quality.

In the phrase grounding task, Florence-2 Large effectively leverages descriptive text cues to identify near-side players with high accuracy in both professional and non-professional settings. Nonetheless, this model struggles significantly with far-side players in the NCAA video, underscoring its susceptibility to diminished visual cues at increased distances and reduced clarity.

Pose estimation using YOLO-Pose demonstrates superior performance in professional match footage, accurately detecting player poses across all court positions with only minor reductions for far-side players. Conversely, its accuracy sharply decreases in lower-quality NCAA footage, entirely missing detections for far-side players. 

The integrated approach, combining Grounding DINO with YOLO-Pose, provides the best overall balance between accurate identification and robust pose estimation. This combined strategy substantially enhances detection and identification accuracy for far-side players compared to using either model independently. More importantly, it can perform phrase grounding, where we can identify and track players across rallies by providing a visual description, which will be extremely invaluable in our tennis annotation tool.

\vspace{1cm}
\textbf{Limitations of GroundingDINO + YOLO-Pose}
\begin{figure}[H]
   \centering
   \includegraphics[width=1\textwidth]{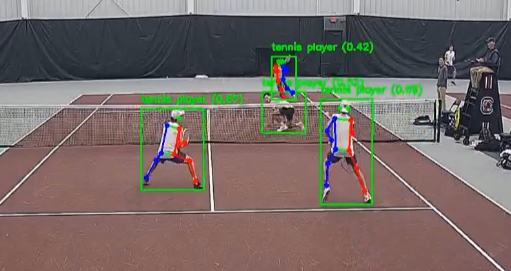}
   \caption{Occlusion of pose estimation model for far players in NCAA video}
   \label{fig:occlusion_pose}
\end{figure}

For several matches such as NCAA doubles, there may be hiccups when dealing with far end (P3, P4) players. There are several potential issues, such as occlusion, when the far player is blocked by the near player shown in this image. This can result in inaccurate (low confidence) pose estimations, or completely missing out pose information for the player in the frame. This can be especially prevalent in lower-quality NCAA matches, where we might not be able to extract information for P4 in up to 16\% of the frames.

In spite of this, a fine-tuned Grounding DINO + YOLO-Pose seems to be the best approach. With an initial annotation of around 10-30 images of the tennis players with visual description with our labelling, we can fine-tune the model and perform accurate inference across all the frames in the tennis video to obtain the bounding boxes of the players.

\clearpage
\textbf{Integration Strategy}
\begin{figure}[H]
   \centering
   \includegraphics[width=1\textwidth]{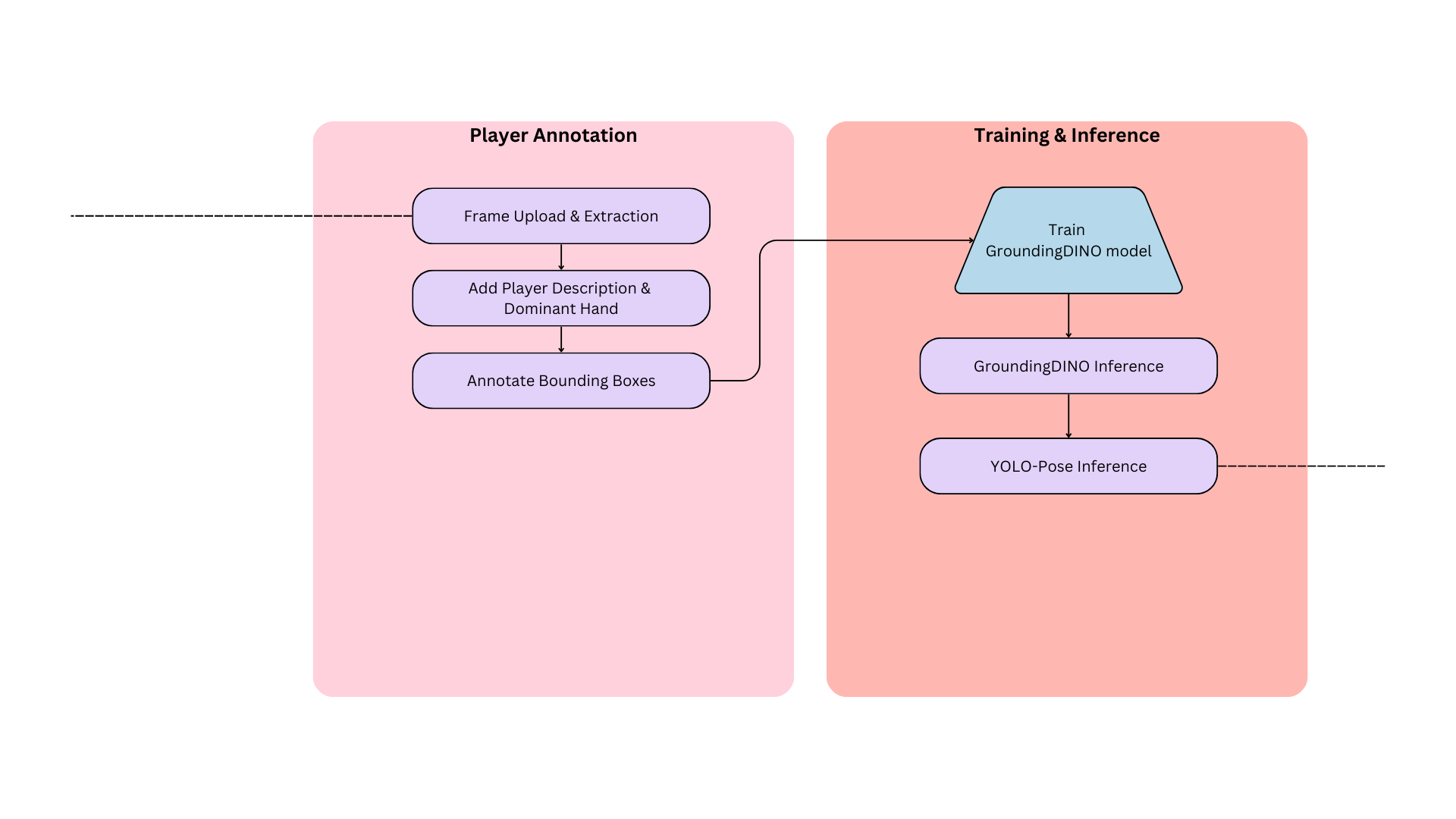}
   \caption{Integration of GroundingDINO \& YOLO-Pose models into the tennis annotation tool}
   \label{fig:integration_strategy}
\end{figure}

To automate the collection of bounding boxes \& pose estimation models, this will involve the integration into the workflow shown in the figure above. In the first phase in the tennis labelling tool, the user will provide the player's visual description (e.g., red-shirt tennis player) for each of the four players and draw bounding boxes for 10-30 frames.

Subsequently, the annotated frames will be used to fine-tune the GroundingDINO model during the model training process. Lastly, inference will be performed across all the frames and with their bounding boxes, the tool will localise each of the four players before applying YOLO-Pose onto them separately, giving us the pose estimations and bounding boxes.
\clearpage 

\subsection{Automated Label Generation}
To reduce the time taken for generation of labels for the shot information, which occurs for every hitting moment in the rally, we can attempt to use a combination of heuristics and machine learning models to automate label generation. These are the following labels for shot generation discussed in section 4.1:
Court Position (Far Deuce, Far Advantage, Near Deuce, Near Advantage),  Shot Side (Forehand, Backhand), Shot Type (Serve, Second-Serve, Return, Volley, Lob, Smash, Swing), Shot Direction (T, B, W, CC, DL, II, IO), Formation (Conventional, I-Formation, Australian) and Outcome (In, Win, Err)

During the training process, to ensure full reproducibility, we have standardised several parameters such as seeding, learning rates, optimisers, batch sizes, etc. The videos for train/val/test sets are split as discussed in Section 3.

Due to the class imbalance for some of the labels, we will be evaluating the model via several metrics such as Macro-average precision, recall and their Area Under the Receiver Operating Characteristic Curve (AUC) scores. These metrics will sufficiently help us determine which model(s) to best integrate into the tennis annotation tool for automated label generation.

\subsubsection{Pose-based Model Architectures}
As discussed in Section 2.2, we can potentially use the extracted player poses to automatically help us generate labels. Using traditional computer vision models for tennis analysis may work well for tennis singles, but due to the added complexity and intricacies of extra players in tennis doubles, computer vision models may not work as well.

Pose-based approaches offer a compelling alternative by focusing on the skeletal structure and joint positions of players rather than pixel-level features. By representing our extracted players as graphs of keypoints (corresponding to joints defined by YOLO-Pose such as shoulders, elbows, wrists), we can sufficiently capture the tennis movements while abstracting away irrelevant visual details. With a new representation of the tennis players' features, our models can then focus directly on the body mechanics that differentiate various shot types, providing a more interpretable feature space.

Additionally, pose-based methods require significantly fewer parameters than conventional convolutional neural networks, making them computationally efficient while potentially improving performance on tasks where body positioning is the primary distinguishing factor (such as forehand/backhand, formation, serve types for tennis doubles).

\subsubsection{Single-Pose GCN Architecture}

\begin{figure}[H]
    \centering
    \includegraphics[width=1\textwidth]{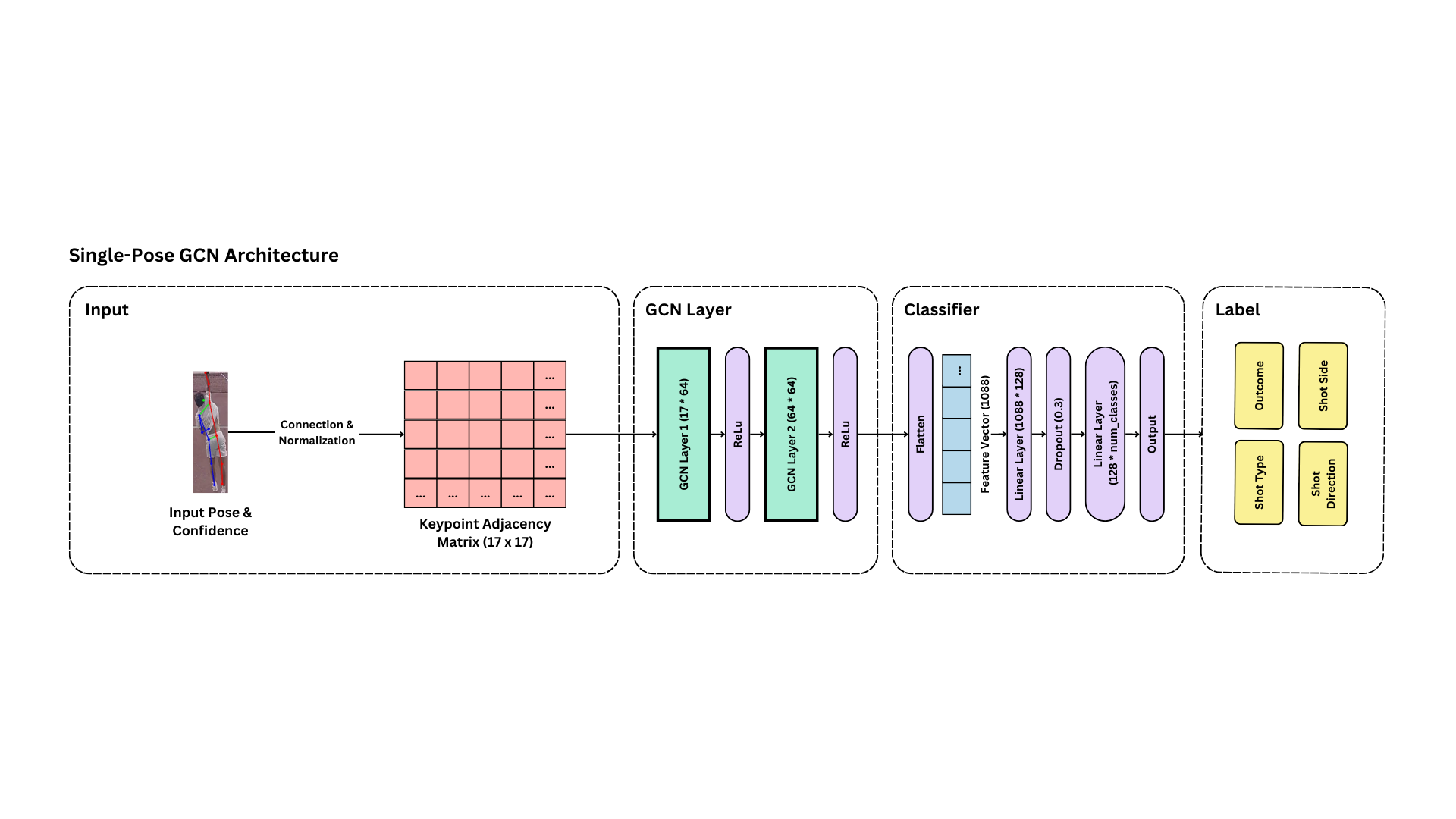}
    \caption{Single-Pose GCN Architecture}
    \label{fig:single_pose_gcn}
\end{figure}

The Single-Pose Graph Convolutional Network (GCN) architecture focuses exclusively on the pose estimations of the hitting player at the moment of ball contact. We represent each player as a graph with 17 nodes corresponding to standard human pose key-points defined by the YOLO-Pose model. Each node contains three features: x-coordinate, y-coordinate, and a confidence score indicating detection reliability. The relationships between joints are encoded in a custom adjacency matrix that mirrors the human skeletal structure, with connections between physically linked joints (e.g., shoulder-to-elbow, elbow-to-wrist) as shown in the keypoint adjacency matrix in Figure~\ref{fig:single_pose_gcn}.

Our model employs multiple graph convolutional layers that propagate information along the skeletal connections, allowing the network to learn complex relationships between joint positions. Each GCN layer applies the transformation:
\begin{equation}
H^{(l+1)} = \sigma(D^{-1/2}AD^{-1/2}H^{(l)}W^{(l)})
\end{equation}
where $A$ is the adjacency matrix, $D$ is the degree matrix, $H^{(l)}$ represents node features at layer $l$, and $W^{(l)}$ contains learnable parameters. After feature extraction, we flatten the node representations and pass them through a classifier to predict shot characteristics. We also employ a class-weighted CrossEntropyLoss function to handle class imbalance, which may be particularly important for labels such as shot type where certain categories (i.e., smash, lob) appear less frequently than others (i.e., serve, swing).

Ideally, the Single-Pose GCN should excel in identifying shot side (forehand/backhand), shot types (serve, volley, etc.), and potentially shot direction (CC/DL/II/IO), which primarily depend on the hitting player's body position and are less influenced by court position or opponent stance.

\vspace{1cm}
\textbf{Single-Pose GCN Results}
\begin{table}[H]
\centering
\small
\begin{tabular}{|>{\raggedright\arraybackslash}p{2.5cm}|c|c|c|c|c|c|c|c|}
\hline
\multirow{2}{*}{\textbf{Label}} & \multicolumn{4}{c|}{\textbf{Professional}} & \multicolumn{4}{c|}{\textbf{NCAA}} \\ \cline{2-9}
 & \textbf{Accuracy} & \textbf{Precision} & \textbf{Recall} & \textbf{AUC} & \textbf{Accuracy} & \textbf{Precision} & \textbf{Recall} & \textbf{AUC} \\ \hline
Side (backhand/forehand) & 67.14\% & 66.44\% & 65.42\% & 68.42\% & 68.14\% & 68.13\% & 68.07\% & 70.22\% \\ \hline
Shot Type (serve vs non-serve) & 52.86\% & 13.21\% & 25.00\% & 42.72\% & 59.29\% & 18.45\% & 28.55\% & 57.22\% \\ \hline
Shot Direction (all) & 23.24\% & 4.98\% & 12.90\% & 55.19\% & 25.05\% & 5.00\% & 14.19\% & 45.86\% \\ \hline
\end{tabular}
\caption{Single-Pose GCN Architecture Results}
\label{tab:single_pose_gcn_results}
\end{table}

Despite being able to perform a non-trivial prediction of the side (forehand/backhand), the overall performance metrics for this Single-Pose GCN model reveal several critical limitations that explain its poor performance, particularly for shot type and shot direction classification.

For side classification, there is a moderate performance of $>65\%$ recall, precision, accuracy and AUC. This classification performs relatively better because the biomechanical differences between forehand and backhand are more directly encoded in static pose data.

However, for shot type and shot direction, there are low recall, precision and AUC values, nearing a substandard worse than random prediction for some videos. This can be explained by several factors, which will also be covered in the Appendix Section:

\begin{itemize}
\item \textbf{Insufficient Temporal Information:} A single static pose cannot capture the dynamic nature of tennis shots, particularly critical for: shot follow-through that determines direction or the sequential movement patterns that differentiate services from other shots
\item \textbf{Missing Contextual Elements:} Currently, racket position/angle is poorly represented in skeletal pose data, court positioning context is absent and ball trajectory information is completely missing
\item \textbf{Pose Data Limitations:} As discussed in Section 4.3.1, for far players, the inaccuracy of the pose extraction models due to various environmental factors (lighting, occlusion) may result in missing information for many of these labels. This can make prediction tasks for even shot side and shot type extremely difficult, where the pose estimations are the only data we are using.
\item \textbf{Insufficient Data:} As discussed in Section 3, generating the initial datasets for label prediction can be extremely time-consuming. Given the course of the project, we only have a total of 8 videos covering just over 2000 samples for all train/val/test sets, making it difficult to train an entire GCN model from scratch, which we cannot leverage on transfer learning due to unique pose information for tennis doubles.
\end{itemize}

\subsubsection{Double-Pose GCN Architecture}
To circumvent some of the limitations, such as limited temporal and partner information, I have explored this Double-Pose GCN architecture in an attempt to provide better model predictions.

\begin{figure}[H]
    \centering
    \includegraphics[width=1\textwidth]{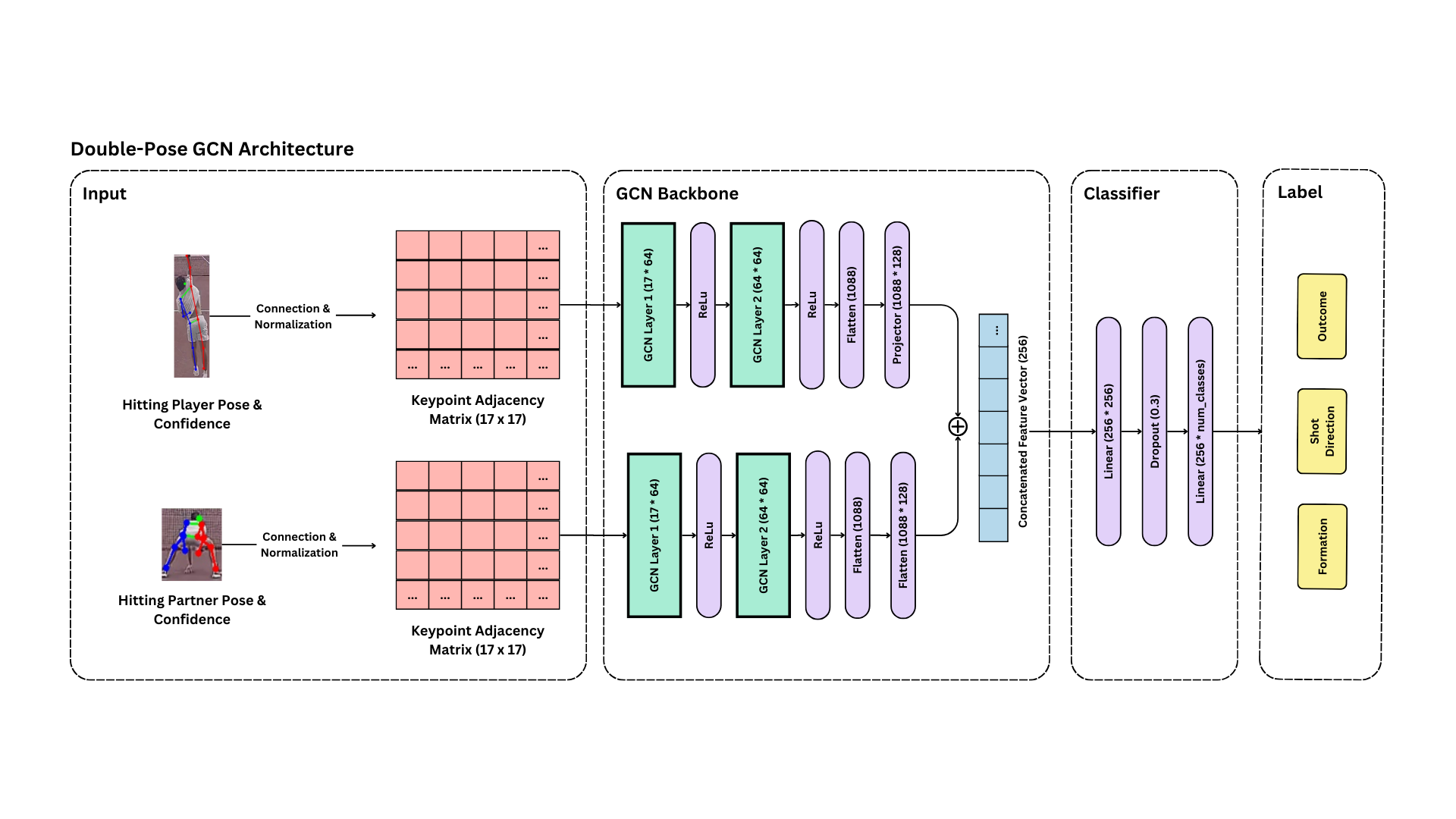}
    \caption{Double-Pose GCN Architecture}
    \label{fig:double_pose_gcn}
\end{figure}

The Double-Pose GCN extends our approach by simultaneously analysing two distinct poses, instead of one shown in the previous architecture, enabling the recognition of complex intricacies in tennis doubles. This architecture consists of two parallel GCN backbones, each processing a different pose. 

Depending on the prediction target, we utilise either the hitting player and partner poses (for formation analysis) or the hitting player's current and future poses (for outcome and direction prediction). Each pose passes through its dedicated GCN backbone, generating a compact feature representation. These representations are then concatenated and processed through a classifier network to produce the final prediction.

This dual-input approach should theoretically enhance our ability to predict team-dependent attributes like formation (conventional, I-formation, Australian) by capturing spatial relationships between partners (i.e., if the partner is crouching down, it is likely I-formation). For temporal predictions such as shot direction and outcome (in, win, err), we leverage poses from subsequent frames (currently set to $n=10$ frames ahead) to model the evolution of the player's stance following the impact frame. This model can potentially learn follow-through mechanics that correlate with shot trajectories and outcomes. By maintaining the graph structure throughout processing, the Double-Pose GCN can preserve important spatial relationships between keypoints while enabling feature sharing across the skeleton.

\vspace{1cm}
\textbf{Double-Pose GCN Results}
\begin{table}[H]
\centering
\small
\begin{tabular}{|p{2.5cm}|c|c|c|c|c|c|c|c|}
\hline
\multirow{2}{*}{\textbf{Label}} & \multicolumn{4}{c|}{\textbf{Professional}} & \multicolumn{4}{c|}{\textbf{NCAA}} \\ \cline{2-9}
 & \textbf{Accuracy} & \textbf{Precision} & \textbf{Recall} & \textbf{AUC} & \textbf{Accuracy} & \textbf{Precision} & \textbf{Recall} & \textbf{AUC} \\ \hline
Shot Direction (all) & 23.24\% & 9.11\% & 17.35\% & 58.75\% & 17.39\% & 4.10\% & 13.09\% & 46.98\% \\ \hline
Formation & 66.67\% & 41.03\% & 41.03\% & 53.58\% & 63.19\% & 44.64\% & 44.64\% & 48.51\% \\ \hline
Outcome & 78.84\% & 26.74\% & 33.33\% & 37.93\% & 70.60\% & 21.90\% & 33.33\% & 54.27\% \\ \hline
\end{tabular}
\caption{Double-Pose GCN Architecture Results}
\label{tab:double_pose_gcn_results}
\end{table}

Similar to the Single-Pose GCN results, the Double-Pose GCN architecture yielded sub-standard predictions. This is also likely due to the similar limitations, such as insufficient data, especially for imbalance classes (formation, outcome), and other contextual or pose data limitations. In addition, with the addition of another player, the size of the GCN backbone has doubled, increasing its overall complexity while using the same number of data points, which might simply be insufficient.

Moving forward, we will be experimenting with architectures that allow us to perform transfer learning, where we can either tap on other existing similar data sources, or pre-trained models.

\subsubsection{CNN-Based Architectures}
While pose-based methods offer efficient representations of player movements, direct image-based approaches provide complementary advantages by capturing comprehensive visual information including court positioning, tennis racket orientation, and subtle body positioning that might not be fully represented in skeletal models. CNNs excel at extracting hierarchical visual features from raw pixel data, allowing our system to recognise patterns that extend from extracted and localised images of hitting players.

Instead of traditionally using the entire video frame for Convolutional Neural Networks (CNNs) based feature extractions and classifications, we will use the localisation method discussed in Section 4.3.1, where we only pass in the image of hitting player(s).

Our image-based architectures also leverage transfer learning from models pre-trained on ImageNet, adapting these powerful feature extractors to tennis-specific classification tasks (Shin et al., 2016). Currently, our self-labelled dataset is still incredibly small, as discussed in Section 3. This may have contributed to a relatively lacklustre performance for our models in the pose-estimation models, where we usually require larger datasets. Therefore, using the CNN approach with pre-trained models can help us overcome the limited size of tennis-specific datasets while maintaining high performance. By implementing both CNN-based and pose-based models, our system can benefit from their complementary strengths, with CNNs capturing rich contextual information and visual cues that might be missed in pose-only representations.

\subsubsection{Single-Image CNN Architecture}

\begin{figure}[H]
    \centering
    \includegraphics[width=1\textwidth]{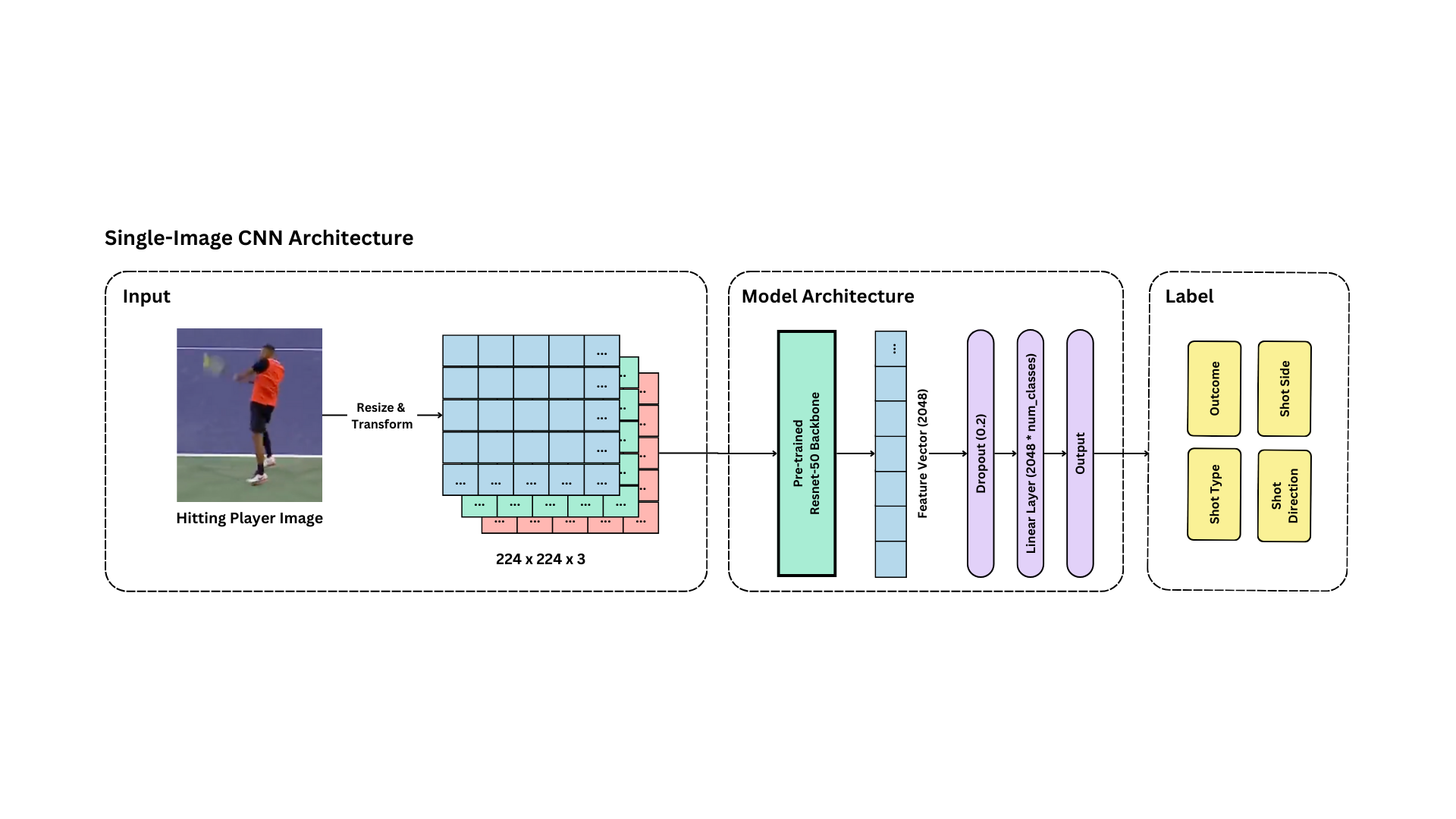}
    \caption{Single-Image CNN Architecture}
    \label{fig:single_image_cnn}
\end{figure}

In this architecture, we first start off by extracting the hitting player's bounding boxes with a margin twice the original size to capture full body positioning and racket movement. Images are subsequently standardised to $224 \times 224$ pixels and augmented during training with random crops, horizontal flips and color jittering to improve generalisability. 

We also employ a class-weighted CrossEntropyLoss function to handle class imbalance, which may be particularly important for labels such as shot type where certain categories (i.e., smash, lob) appear less frequently than others (i.e., serve, swing). The model is also trained using an AdamW optimiser with differential learning rates–lower for pre-trained backbone (LR/10) and higher for classifier head (LR=0.0001) to enable efficient knowledge transfer and adaptability to the tennis-specific features present in the image.

Similarly, this architecture should perform especially well for classifying labels that can be determined in a single frame, such as shot side, shot type, and potentially shot direction, where the individual player appearance provides sufficient information.

\clearpage
\textbf{Single-Image CNN Results}
\begin{table}[H]
\centering
\small
\begin{tabular}{|>{\raggedright\arraybackslash}p{2.5cm}|c|c|c|c|c|c|c|c|}
\hline
\multirow{2}{*}{\textbf{Label}} & \multicolumn{4}{c|}{\textbf{Professional}} & \multicolumn{4}{c|}{\textbf{NCAA}} \\ \cline{2-9}
 & \textbf{Accuracy} & \textbf{Precision} & \textbf{Recall} & \textbf{AUC} & \textbf{Accuracy} & \textbf{Precision} & \textbf{Recall} & \textbf{AUC} \\ \hline
Side (backhand/forehand) & 75.00\% & 73.24\% & 68.16\% & 75.75\% & 70.06\% & 66.72\% & 67.37\% & 71.82\% \\ \hline
Shot Type (serve vs non-serve) & 100.00\% & 100.00\% & 100.00\% & 100.00\% & 90.40\% & 90.98\% & 87.78\% & 94.88\% \\ \hline
Shot Type (all) & 80.68\% & 78.52\% & 80.68\% & 86.65\% & 75.71\% & 70.22\% & 55.48\% & 83.11\% \\ \hline
Shot Direction (all) & 39.77\% & 43.09\% & 39.77\% & 74.05\% & 34.09\% & 30.43\% & 24.68\% & 73.75\% \\ \hline
Shot Direction (T/B/W/ Cross/Straight) & 57.95\% & 44.18\% & 55.97\% & 76.08\% & 44.63\% & 38.76\% & 38.20\% & 77.61\% \\ \hline
Shot Direction (Cross/Straight) & 69.32\% & 68.79\% & 69.34\% & 72.28\% & 71.19\% & 67.42\% & 68.16\% & 74.40\% \\ \hline
Formation & 93.18\% & 72.86\% & 76.98\% & 97.79\% & 86.44\% & 72.16\% & 77.32\% & 95.48\% \\ \hline
Outcome & 69.32\% & 30.39\% & 30.21\% & 56.32\% & 55.93\% & 45.73\% & 47.48\% & 61.80\% \\ \hline
\end{tabular}
\caption{Single-Image CNN Architecture Results}
\label{tab:single_image_cnn_results}
\end{table}

Compared to the GCN models, the Single-Frame CNN demonstrated strong performance across several prediction tasks:
\begin{itemize}
\item \textbf{Shot Type Classification:} Achieved perfect accuracy (100\%) for distinguishing between serves and non-serves on professional matches, and excellent performance (90.40\% accuracy) on NCAA matches. In addition, for predicting other shot types (volley, swing, lob), it boasts impressive AUC scores of $>80\%$ and relatively high precision and recall, especially for professional matches.
\item \textbf{Formation Prediction:} Despite using only a single player's image, the model achieved 93.18\% accuracy on professional matches and 86.44\% on NCAA matches, indicating that player positioning alone contains strong signals for formation.
\item \textbf{Side Detection:} The model effectively distinguished between forehand and backhand shots with 75.00\% accuracy on professional matches and 70.06\% on NCAA matches.
\end{itemize}

The better performance compared to our GCN models can be explained by the use of a pre-trained ResNet-50 backbone. This backbone has been pre-trained for other tasks, and by using the concept of model distillation, we can easily alleviate some issues with the limited number of data points that we have, especially for labels with major imbalances, such as formation and shot type.

Despite the decent performances for Shot Type, Side and Formation, classification for Shot Direction and Outcome falls short, with much lower scores of precision and recall across the board. The result is expected, as outcome and formation both depends on their partners or temporal information during the rally, which explains the inability for our Single-Image CNN architecture to accurately pick up this information.

\subsubsection{Double-Image CNN Architecture}

\begin{figure}[H]
    \centering
    \includegraphics[width=1\textwidth]{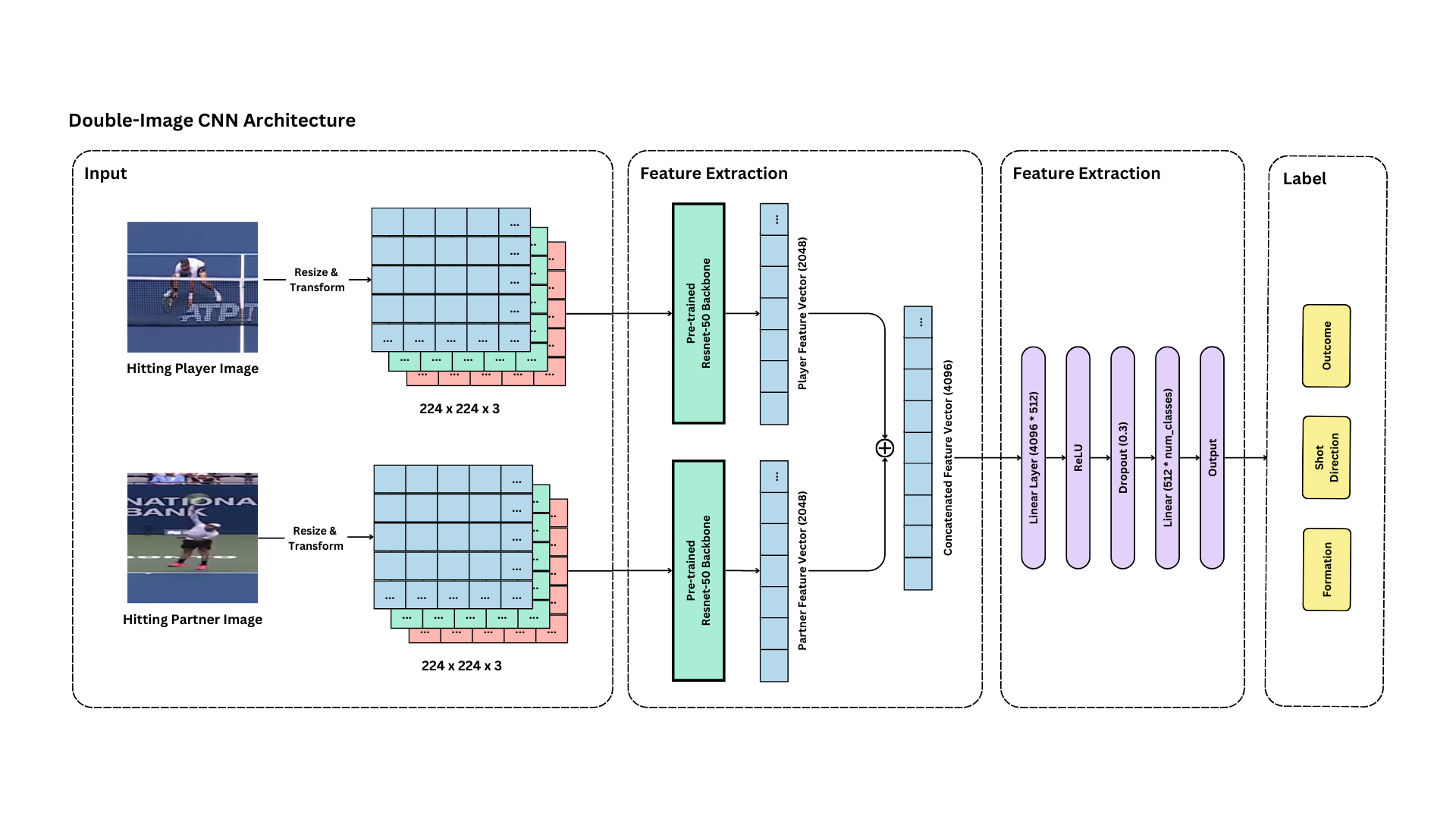}
    \caption{Double-Image CNN Architecture}
    \label{fig:double_image_cnn}
\end{figure}

The Double-Frame CNN architecture extends our approach (of both single-image CNN and double-pose GCN) by simultaneously analysing two images, enabling more complex pattern recognition for formation and outcome-based predictions. This model consists of dual ResNet50 backbones, each processing a different image to extract the respective player features. Depending on the prediction target, we utilise either the hitting player and partner images for formation or the hitting player's current and future frames for outcome and shot direction prediction, similarly, with $n=10$ frames in the future. Each image passes through its dedicated backbone, generating a 2048-dimensional feature vector. These features are concatenated and processed through a multi-layer classifier network with 512 hidden units, ReLU activation, and dropout regularization (rate=0.3) to produce the final prediction.

This architecture should be able to enhance our ability to predict team-dependent attributes like formation with both the hitting player and partner's images. The temporal aspects of the shot should also be extracted with a future frame, thereby helping in identifying the shot outcome and direction. Training the model follows a similar approach to the single-frame model but with slightly more layers to effectively concatenate and represent both the frame's features. 

\textbf{Double-Image CNN Results}
\begin{table}[H]
\centering
\small
\begin{tabular}{|>{\raggedright\arraybackslash}p{2.5cm}|c|c|c|c|c|c|c|c|}
\hline
\multirow{2}{*}{\textbf{Label}} & \multicolumn{4}{c|}{\textbf{Professional}} & \multicolumn{4}{c|}{\textbf{NCAA}} \\ \cline{2-9}
 & \textbf{Accuracy} & \textbf{Precision} & \textbf{Recall} & \textbf{AUC} & \textbf{Accuracy} & \textbf{Precision} & \textbf{Recall} & \textbf{AUC} \\ \hline
Shot Direction (all) & 42.05\% & 15.49\% & 14.54\% & 72.43\% & 31.07\% & 29.38\% & 22.92\% & 70.98\% \\ \hline
Shot Direction (T/B/W/ Cross/Straight) & 59.09\% & 31.24\% & 28.38\% & 79.80\% & 44.63\% & 38.88\% & 38.42\% & 34.16\% \\ \hline
Shot Direction (Cross/Straight) & 68.18\% & 65.83\% & 58.76\% & 61.86\% & 72.32\% & 50.44\% & 59.3\% & 68.13\% \\ \hline
Formation & 94.28\% & 82.42\% & 88.10\% & 99.25\% & 87.57\% & 73.83\% & 88.74\% & 96.92\% \\ \hline
Outcome & 69.32\% & 36.56\% & 41.62\% & 69.41\% & 64.14\% & 45.73\% & 47.48\% & 65.56\% \\ \hline
\end{tabular}
\caption{Double-Image CNN Architecture Results}
\label{tab:double_image_cnn_results}
\end{table}

A key difference between the Single-Image CNN architecture from the Double-Image CNN architecture is the integration of temporal and partner information. As evident, this results in slightly better performance metrics for both formation and outcome, which requires this information. 

However, surprisingly, prediction of the shot direction has a poorer performance for this architecture. This may be explained by several factors, including that using $n=10$ frames worth of temporal information may be unsuitable, or the use of the player's image may not be as useful as information obtained from the ball's trajectory, which might be more suitable for determining shot direction.

\subsubsection{Overall AUC Comparison}

\begin{table}[H]
\centering
\small
\begin{tabular}{|>{\centering\arraybackslash}p{3cm}|c|c|c|c|}
\hline
\multicolumn{5}{|c|}{\textbf{Professional}} \\
\hline
\textbf{Label} & \textbf{Single-Pose GCN} & \textbf{Double-Pose GCN} & \textbf{Single-Image CNN} & \textbf{Double-Image CNN} \\
\hline
Side (backhand/forehand) & 68.42\% & - & \textbf{75.75\%} & - \\
\hline
Shot Type (serve vs non-serve) & 42.72\% & - & \textbf{100.00\%} & - \\
\hline
Shot Type (all) & - & - & \textbf{86.65\%} & - \\
\hline
Shot Direction (all) & 55.19\% & 58.75\% & \textbf{74.05\%} & 72.43\% \\
\hline
Shot Direction (T/B/W/ Cross/Straight) & - & - & 76.08\% & \textbf{79.80\%} \\
\hline
Shot Direction (Cross/Straight) & - & - & \textbf{72.28\%} & 61.86\% \\
\hline
Formation & - & 53.58\% & 97.79\% & \textbf{99.25\%} \\
\hline
Outcome & - & 37.93\% & 56.32\% & \textbf{69.41\%} \\
\hline
\end{tabular}
\caption{Overall comparison of model AUC for Professional Tennis Videos}
\label{tab:professional_auc_comparison}
\end{table}

\begin{table}[H]
\centering
\small
\begin{tabular}{|>{\centering\arraybackslash}p{3cm}|c|c|c|c|}
\hline
\multicolumn{5}{|c|}{\textbf{NCAA}} \\
\hline
\textbf{Label} & \textbf{Single-Pose GCN} & \textbf{Double-Pose GCN} & \textbf{Single-Image CNN} & \textbf{Double-Image CNN} \\
\hline
Side (backhand/forehand) & 70.22\% & - & \textbf{71.82\%} & - \\
\hline
Shot Type (serve vs non-serve) & 57.22\% & - & \textbf{94.88\%} & - \\
\hline
Shot Type (all) & - & - & \textbf{83.11\%} & - \\
\hline
Shot Direction (all) & 45.86\% & 46.98\% & \textbf{73.75\%} & 70.98\% \\
\hline
Shot Direction (T/B/W/ Cross/Straight) & - & - & \textbf{77.61\%} & 34.16\% \\
\hline
Shot Direction (Cross/Straight) & - & - & \textbf{74.40\%} & 68.13\% \\
\hline
Formation & - & 48.51\% & 95.48\% & \textbf{96.92\%} \\
\hline
Outcome & - & 54.27\% & 61.80\% & \textbf{65.56\%} \\
\hline
\end{tabular}
\caption{Overall comparison of model AUC for NCAA Tennis Videos}
\label{tab:ncaa_auc_comparison}
\end{table}

\subsubsection{Integration into Tennis Labelling Tool}

\begin{figure}[H]
    \centering
    \includegraphics[width=0.9\textwidth]{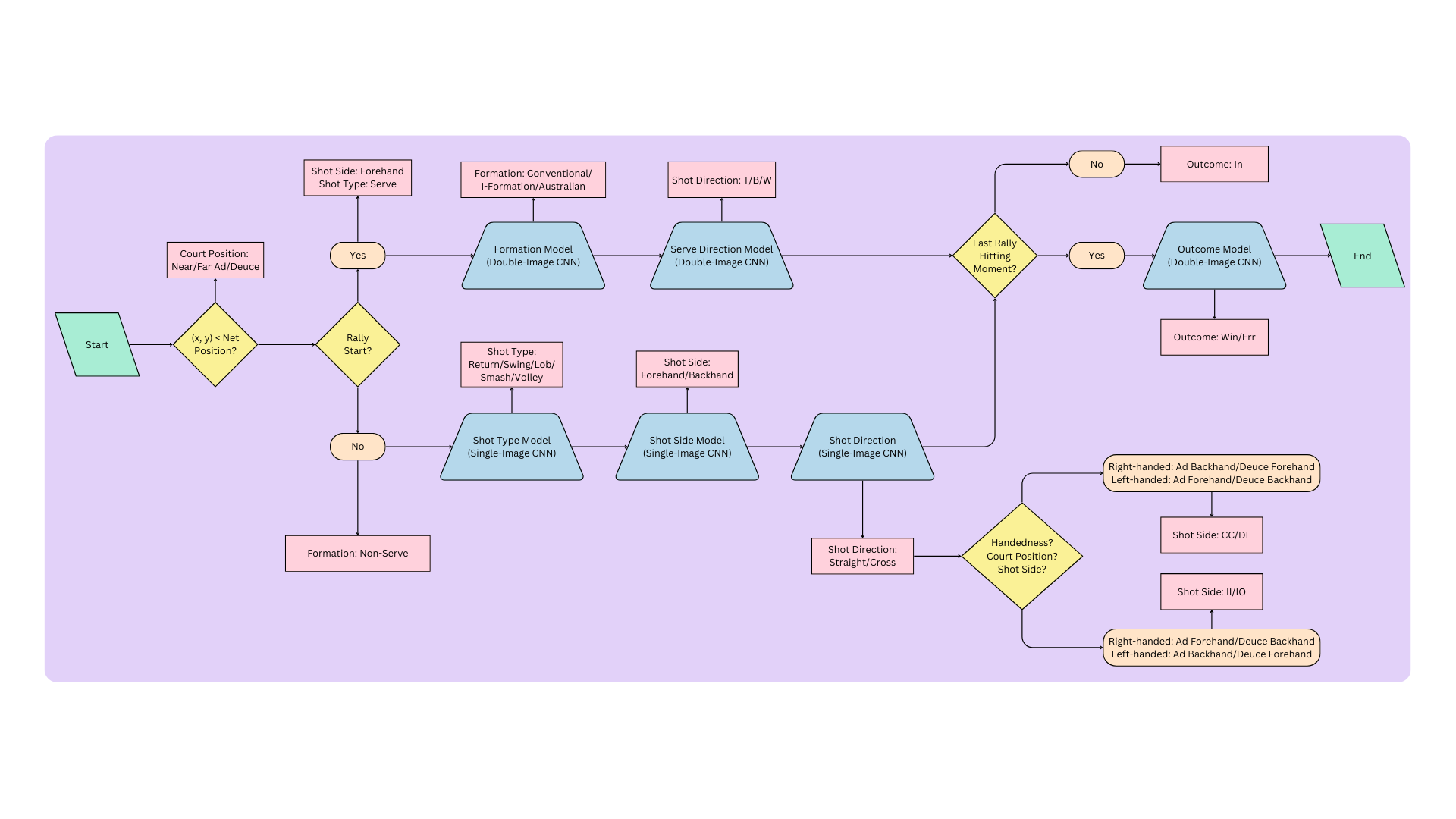}
    \caption{Integration Workflow}
    \label{fig:integration_workflow}
\end{figure}

Given the results shown in Tables \ref{tab:professional_auc_comparison} and \ref{tab:ncaa_auc_comparison} above, we will be integrating the CNN-based architectures for label generation, given their superior performances. This is the following workflow:
\begin{enumerate}
\item Determine court position with the hitting player's and net's positions
\item Given a serve, determine the formation and serve direction
\item Given a non-serve, determine the shot type, shot side and shot direction
\item Lastly, determine the outcome if it is the last shot in the rally
\end{enumerate}

We will use the Single-Image CNN architecture for Shot Side, Shot Type and Shot Direction predictions and Double-Image CNN architecture for Formation and Outcome predictions. These models will be pre-trained and integrated into the tennis annotation tool.

\chapter{Conclusion}
\section{Key Contributions}
This project has successfully developed and implemented a comprehensive video-based analytics framework tailored specifically to tennis doubles as outlined in the project objectives under Section 1.3. The key contributions of the project can be summarised as follows: (1) The creation of a standardised annotation framework and its corresponding initial dataset to establish systematic data collection for tennis doubles, (2) The development of a user-friendly, integrated annotation tool to streamline tennis doubles analysis that significantly reduces manual annotation time through its innovative features and lastly (3) The novel application and evaluation of suitable model architectures for tennis doubles through a systematic comparison of pose-based GCN versus CNN-based image classification models.

\section{Future Work}
Building onto the current capabilities of our tennis doubles analytics framework, our future efforts will prioritise the enhancement of the tennis annotation tool features, significant model improvements and the expansion of implicit/explicit data features to be collected.

\vspace{0.5cm}
\textbf{Enhanced Tennis Annotation Tool Features}
\begin{itemize}
\item \textbf{Retraining Capabilities:} Integrating seamless model retraining directly within the tennis annotation tool will be one of the significant priorities. As additional labeled data accumulates through regular annotation processes, these new data points will continuously feed into the existing models, allowing incremental fine-tuning. This iterative retraining workflow ensures the models remain updated and progressively improve their predictive performances, maintaining high performance as more annotated data becomes available. This will likely be integrated into our frontend, where after labelling the new data points, we can choose to retrain selected models and potentially identify the models' new performance metrics and if there are any data drifts.

\item \textbf{Interactive Feedback Loop:} Establishing an interactive mechanism for annotators to provide real-time feedback on model predictions will allow the annotation tool to dynamically highlight areas of uncertainty or potential errors. This allows for a human-in-the-loop approach that will significantly enhance annotation accuracy and efficiency. 

\item \textbf{Extension to other sports:} Given that we have a set framework and workflow for tennis doubles, we might be able to easily extend to similar sports such as tennis singles, or even doubles sports such as table tennis doubles. The framework that we built is meant to be easily modifiable and extendible to other domains.
\end{itemize}

\vspace{0.5cm}
\textbf{Model Improvement and Advanced Architectures}
\begin{itemize}
\item \textbf{Multimodal Architectures:} With increased data collection, exploring multimodal architectures that integrate visual data (player poses, racket and ball positions) with additional modalities (e.g., audio signals, player movement trajectories, and temporal dynamics) could enhance predictive power and contextual understanding. Employing transformer-based multimodal architectures, as demonstrated by Liu et al (2025), could particularly benefit shot type, shot direction, and outcome prediction.

\item \textbf{Integration of Ball and Racket Tracking:} Our current prediction framework operates in isolated silos leveraging multiple models followed by rule-based heuristics for each prediction task (i.e., determining shot side, formation, serve/shot directions). One future direction could be creating a singular, integrated model consisting of multiple branches -- ball, racket, pose, player positioning, where information from each branch can enhance the performance for our final prediction task. Ultimately, with a sufficiently large feature set and data points, our model can assimilate information from all the branches to provide a singular classification of the event type (e.g., near\_ad\_forehand\_serve\_b\_i-formation\_in). For further research, we may explore models like TrackNet (Huang et al., 2019), which propose innovative trajectory prediction frameworks to enhance the accuracy and robustness of automated annotations.
\end{itemize}

\section{Final Thoughts}
My journey of developing this tennis doubles analytics framework and tool has provided deep insights into both the opportunities and complexities within the field of sports analytics. Through addressing the unique analytical challenges posed by tennis doubles, my project reflected the incredible challenges faced when collecting accurate and standardised datasets for sports analytics. Nonetheless, it has also shown a transformative potential of integrating state-of-the-art machine learning algorithms into these intuitive annotation tools to greatly reduce manual efforts of obtaining such time-consuming and intricate datasets.

While substantial progress has been made, especially in automating traditionally manual annotation tasks, it is clear that continued innovation and iterative improvement are essential. Moving forward, incorporating richer multimodal data and advanced temporal modeling will be crucial in further bridging the gap between analytical capabilities and practical sports strategy.

Ultimately, the framework developed during the course of my project sets a solid foundation for future research, innovation, and practical applications, highlighting the increasingly important role data science and analytics play in enhancing sports performance and strategy.

\addcontentsline{toc}{chapter}{References}

\appendix
\chapter{Appendix}
\section{Previous Annotation Tool}
During the early stages of this project, we have developed a rudimentary annotation tool to create the initial dataset required for exploratory analysis and preliminary model training. This first-generation tool served as a foundational step towards understanding the specific annotation challenges posed by tennis doubles matches.

\begin{figure}[H]
\centering
\includegraphics[width=1\textwidth]{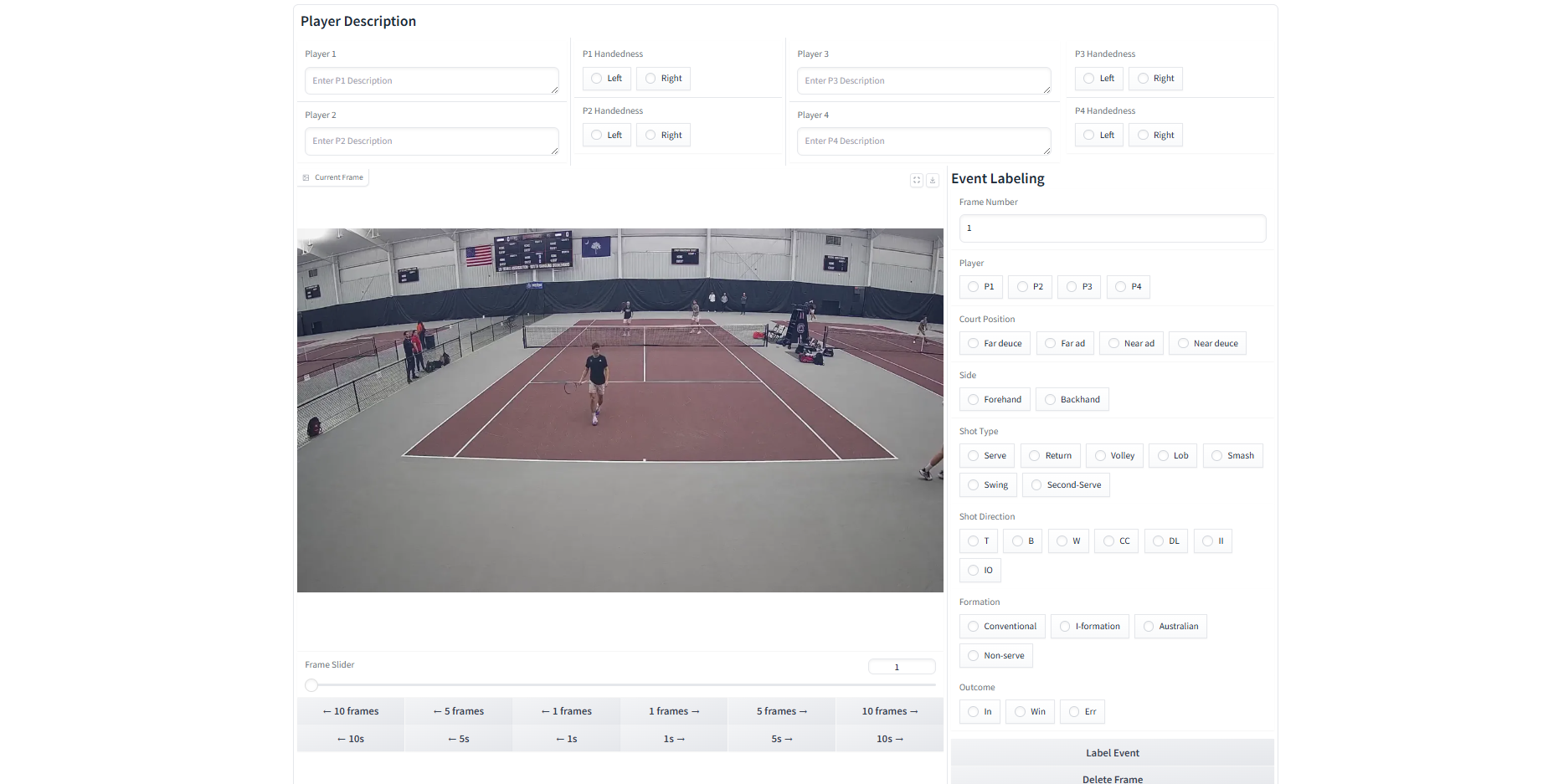}
\caption{Preliminary annotation tool interface}
\label{fig}
\end{figure}

The preliminary tool was significantly limited in scope and functionality compared to the comprehensive framework presented in the main report. It possessed only basic video navigation capabilities and simple annotation features, allowing for manual marking of player positions, shot types, and basic rally events. The interface followed a straightforward design with minimal controls for frame navigation and label selection. There are no integration with ML models, absence of rally analysis, no training and inference infrastructure with limited annotation schema.

\section{Exploration of 3D Pose Estimation}

While the main framework focused primarily on 2D pose estimation and bounding box detection, we have explored the potential integration of 3D pose estimation into the tennis doubles analysis pipeline. Three-dimensional pose data could potentially provide richer representations of player movements, enabling more accurate shot classification and strategic pattern recognition.

\subsection{From 2D to 3D Pose via MotionAGFormer}

\subsubsection{Single Player Pose Estimation}
In the methodology proposed by Mehraban et al. (2023), 2D poses are firstly extracted from the detected player with the highest confidence score via a pre-trained YOLOv3 model. Subsequently, the MotionAGFormer pipeline is used to transform these 2D poses into 3D representations through the following process:

\begin{enumerate}
    \item \textbf{Feature Embedding}
    \begin{itemize}
        \item Given a 2D Pose Sequence $X \in \mathbb{R}^{T\times J\times 3}$ of $T$ frames, $J$ joints \& (x, y coordinates, confidence score)
        \item Create an initial feature map
        \item Add a spatial positional embedding to maintain joint relationships
    \end{itemize}
    
    \item \textbf{AGFormer Block Processing}
    \begin{itemize}
        \item Transformer stream: Uses spatial \& temporal Multi-Head Self-Attention (MHSA) to capture global relationships between all joints
        \item GCNFormer stream: Uses spatial \& temporal Graph Convolutional Networks to focus on local joint relationships
    \end{itemize}
    
    \item \textbf{Post-processing}
    \begin{itemize}
        \item Map to a higher dimension and output a 3D pose sequence $(\hat{P} \in \mathbb{R}^{T\times J\times 3})$
        \item Trained using a comprehensive loss function comprising:
        \begin{itemize}
            \item Position Loss: Measures accuracy of 3D positions
            \item Velocity Loss: Ensures smooth motion between frames
        \end{itemize}
    \end{itemize}
\end{enumerate}

\begin{figure}[H]
    \centering
    \includegraphics[width=1\textwidth]{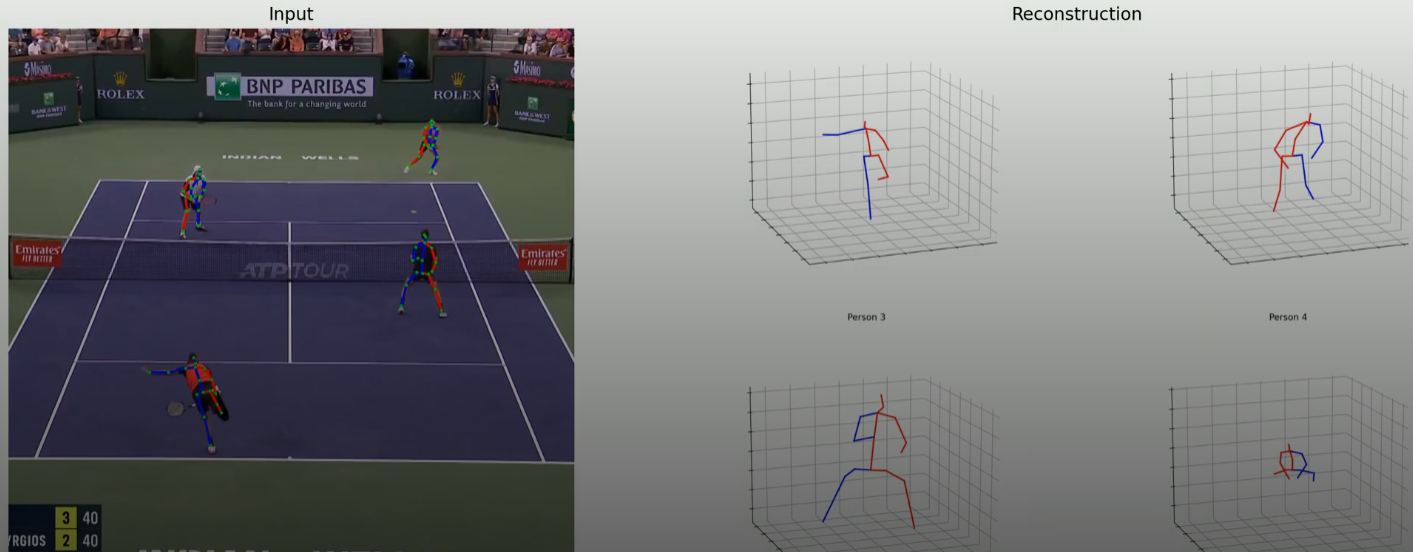}
    \caption{Multiple Player Pose Estimation via MotionAGFormer}
    \label{fig:single_player_3d}
\end{figure}

When applied to a tennis doubles match, this approach successfully tracked individual players' 3D motions throughout rallies. However, the original implementation was limited to processing a single player, necessitating modifications to capture all four players simultaneously.

Our experimental approach leveraged MotionAGFormer's architecture while modifying the input pipeline to handle multiple concurrent pose sequences. This adaptation required restructuring the input tensor from a single pose sequence $X \in \mathbb{R}^{T\times J\times 3}$ to a multi-person sequence $X_m \in \mathbb{R}^{T\times 4J\times 3}$, where each frame contains joint coordinates and confidence scores for all four players on court.

However, due to several limitations (such as an extremely high computational overhead, long processing time and distorted 3D mapping for far players), 3D pose estimation is not yet integrated into our annotation tool. In the future, with more optimisations and newer models, it will be more feasible to integrate 3D pose estimation into our framework.
\clearpage 

\section{Model Training Process \& Hyperparameter Selection}
\textbf{Experimentation Code:} \url{https://github.com/jiaawe/tennis-prediction}

\begin{figure}[H]
\centering
\includegraphics[width=1\textwidth]{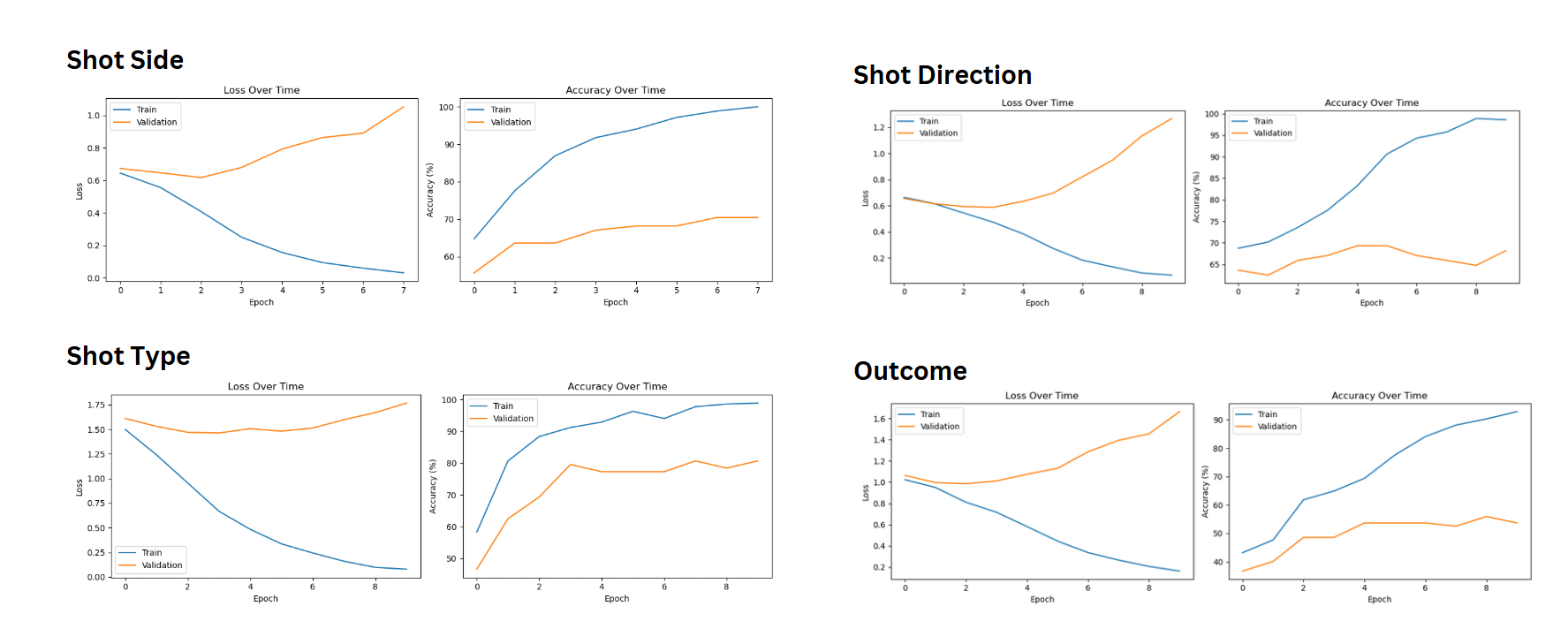}
\caption{Training and validation metrics for Single-Image CNN architecture}
\label{training_progress_cnn}
\end{figure}

\begin{figure}[H]
\centering
\includegraphics[width=1\textwidth]{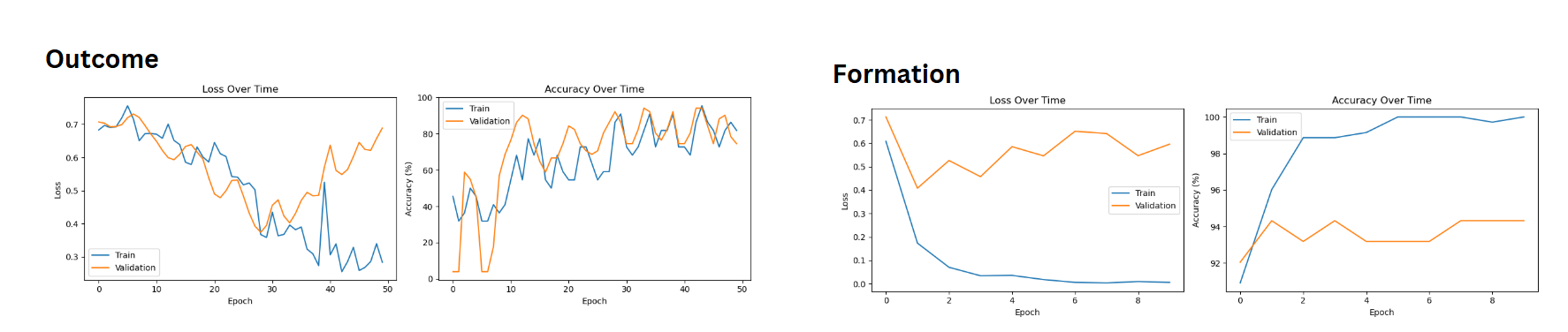}
\caption{Training and validation metrics for Double-Image CNN architecture}
\label{training_progress_double_cnn}
\end{figure}

The training processes for the Single-Image CNN and Double-Image CNN architectures were carefully monitored to determine optimal hyperparameters and prevent overfitting. Figures \ref{training_progress_cnn} and \ref{training_progress_double_cnn} illustrate the loss and accuracy trajectories for different prediction tasks across training epochs.
The training curves revealed several critical insights that informed our final implementation. For the Single-Image CNN, we observed that most tasks achieved optimal validation performance within 10 epochs, with diminishing returns or potential overfitting thereafter. This rapid convergence can be attributed to our transfer learning approach, which leveraged pre-trained ResNet-50 weights and required only focussed on fine-tuning the upper layers to adapt to tennis-specific features. Based on these observations, we implemented an early stopping mechanism with a patience of 20 epochs (monitoring validation loss) to automatically terminate training when no further improvements were detected.

For hyper-parameter selection, we employed a differential learning rate strategy, applying a lower learning rate (1e-5) to the pre-trained backbone and a 10-fold higher rate (1e-4) to the classifier head, which proved crucial for effective knowledge transfer. This approach, combined with the utilisation of pre-trained weights, contributed significantly to the CNN architectures outperforming their GCN counterparts, which required training from scratch on our limited dataset.

\begin{thebibliography}{99}

\bibitem{alshami2023}
AlShami, A., Boult, T., \& Kalita, J. (2023). Pose2Trajectory: Using transformers on body pose to predict tennis player's trajectory. \textit{Journal of Visual Communication and Image Representation}, 97, 103954. https://doi.org/10.1016/j.jvcir.2023.103954

\bibitem{dartfish2025}
Dartfish. (2025). Dartfish video analysis software. https://www.dartfish.com/

\bibitem{do2024}
Do, J., \& Kim, M. (2024). SkateFormer: Skeletal-Temporal Transformer for Human Action Recognition. \textit{arXiv preprint arXiv:2403.09508}. http://arxiv.org/abs/2403.09508

\bibitem{dutta2019}
Dutta, A., \& Zisserman, A. (2019). The VIA annotation software for images, audio and video. \textit{Proceedings of the 27th ACM International Conference on Multimedia}. https://www.robots.ox.ac.uk/\~vgg/software/via/

\bibitem{encord2025}
Encord. (2025). Data-centric computer vision platform. https://encord.com/

\bibitem{gao2024}
Gao, M., \& Ju, B. (2024). Attention-enhanced gated recurrent unit for action recognition in tennis. \textit{PeerJ Computer Science}, 10, e1804. https://doi.org/10.7717/peerj-cs.1804

\bibitem{huang2019}
Huang, Y.-C., Liao, I.-N., Chen, C.-H., İk, T.-U., \& Peng, W.-C. (2019). TrackNet: A Deep Learning Network for Tracking High-speed and Tiny Objects in Sports Applications. \textit{arXiv preprint arXiv:1907.03698}. https://arxiv.org/abs/1907.03698

\bibitem{humansignal2025}
HumanSignal. (2025). Label Studio: Open source data labeling platform. https://labelstud.io/

\bibitem{intel2025}
Intel. (2025). Computer Vision Annotation Tool (CVAT). OpenCV. https://www.cvat.ai/

\bibitem{jiang2020}
Jiang, K., Izadi, M., Liu, Z., \& Dong, J. S. (2020). Deep Learning Application in Broadcast Tennis Video Annotation. \textit{2020 25th International Conference on Engineering of Complex Computer Systems (ICECCS)}, Singapore, 53-62. doi: 10.1109/ICECCS51672.2020.00014

\bibitem{khanam2024}
Khanam, R., \& Hussain, M. (2024). YOLOv11: An Overview of the Key Architectural Enhancements. \textit{arXiv preprint arXiv:2410.17725}. http://arxiv.org/abs/2410.17725

\bibitem{kinovea2023}
Kinovea. (2023). Kinovea: Video analysis for sports. https://www.kinovea.org/

\bibitem{lin2021}
Lin, J., \& Lee, G. H. (2021). Multi-View Multi-Person 3D Pose Estimation with Plane Sweep Stereo. \textit{arXiv preprint arXiv:2104.02273}. http://arxiv.org/abs/2104.02273

\bibitem{liu2024}
Liu, S., Zeng, Z., Ren, T., Li, F., Zhang, H., Yang, J., Jiang, Q., Li, C., Yang, J., Su, H., Zhu, J., \& Zhang, L. (2024). Grounding DINO: Marrying DINO with Grounded Pre-Training for Open-Set Object Detection. \textit{arXiv preprint arXiv:2303.05499}. http://arxiv.org/abs/2303.05499

\bibitem{liu2025}
Liu, Z., Dong, C., Chen, J. W., Jiang, A. M. J., Chen, G., Shaikh, A. F., Dong, T. Y., Wang, C., Jiang, K., \& Dong, J. S. (2025). Analyzing the Formation Strategy in Tennis Doubles Game. \textit{SN Computer Science}, 6(2). https://doi.org/10.1007/s42979-024-03598-3

\bibitem{liu2024f3set}
Liu, Z., Jiang, K., Ma, M., Hou, Z., Lin, Y., \& Dong, J. S. (2024, October 4). $F^3Set$: Towards Analyzing Fast, Frequent, and Fine-grained Events from Videos. \textit{The Thirteenth International Conference on Learning Representations}. https://openreview.net/forum?id=vlg5WRKHxh

\bibitem{maji2022}
Maji, D., Nagori, S., Mathew, M., \& Poddar, D. (2022). YOLO-Pose: Enhancing YOLO for Multi Person Pose Estimation Using Object Keypoint Similarity Loss. \textit{arXiv preprint arXiv:2204.06806}. http://arxiv.org/abs/2204.06806

\bibitem{mehraban2023}
Mehraban, S., Adeli, V., \& Taati, B. (2023). MotionAGFormer: Enhancing 3D Human Pose Estimation with a Transformer-GCNFormer Network. \textit{arXiv preprint arXiv:2310.16288}. http://arxiv.org/abs/2310.16288

\bibitem{shin2016}
Shin, H.-C., Roth, H. R., Gao, M., Lu, L., Xu, Z., Nogues, I., Yao, J., Mollura, D., \& Summers, R. M. (2016). Deep Convolutional Neural Networks for Computer-Aided Detection: CNN Architectures, Dataset Characteristics and Transfer Learning. \textit{IEEE Transactions on Medical Imaging}, 35(5), 1285–1298. https://doi.org/10.1109/TMI.2016.2528162

\bibitem{swingvision2025}
SwingVision. (2025). AI tennis app for video analysis and shot tracking. https://swing.vision/

\bibitem{v7labs2025}
V7 Labs. (2025). AI-powered image and video annotation software. https://www.v7labs.com/

\bibitem{wang2024}
Wang, C. Y., Lai, K. G., Huang, H. C., \& Lin, W. T. (2024). Tennis player actions dataset for human pose estimation. \textit{Data in Brief}, 55, 110665. doi: 10.1016/j.dib.2024.110665

\bibitem{wang2024hulk}
Wang, Y., Wu, Y., Tang, S., He, W., Guo, X., Zhu, F., Bai, L., Zhao, R., Wu, J., He, T., \& Ouyang, W. (2024). Hulk: A Universal Knowledge Translator for Human-Centric Tasks. \textit{arXiv preprint arXiv:2312.01697}. http://arxiv.org/abs/2312.01697

\bibitem{wojke2017}
Wojke, N., Bewley, A., \& Paulus, D. (2017). Simple Online and Realtime Tracking with a Deep Association Metric. \textit{arXiv preprint arXiv:1703.07402}. http://arxiv.org/abs/1703.07402

\bibitem{xiao2023}
Xiao, B., Wu, H., Xu, W., Dai, X., Hu, H., Lu, Y., Zeng, M., Liu, C., \& Yuan, L. (2023). Florence-2: Advancing a Unified Representation for a Variety of Vision Tasks. \textit{arXiv preprint arXiv:2311.06242}. http://arxiv.org/abs/2311.06242

\bibitem{yan2018}
Yan, S., Xiong, Y., \& Lin, D. (2018). Spatial Temporal Graph Convolutional Networks for Skeleton-Based Action Recognition. \textit{arXiv preprint arXiv:1801.07455}. http://arxiv.org/abs/1801.07455

\bibitem{dong2023sports}
Dong, J. S., Jiang, K., Liu, Z., Dong, C., Hou, Z., Hundal, R. S., Guo, J., \& Lin, Y. (2023).
Sports analytics using probabilistic model checking and deep learning.
In \textit{2023 27th International Conference on Engineering of Complex Computer Systems (ICECCS)} (pp. 7--11). IEEE.

\bibitem{liu2023insight}
Liu, Z., Jiang, K., Hou, Z., Lin, Y., \& Dong, J. S. (2023).
Insight analysis for tennis strategy and tactics.
In \textit{2023 IEEE International Conference on Data Mining (ICDM)} (pp. 1169--1174). IEEE.

\bibitem{liu2023recognizing}
Liu, Z., Guo, J., Wang, M., Wang, R., Jiang, K., \& Dong, J. S. (2023).
Recognizing a sequence of events from tennis video clips: addressing timestep identification and subtle class differences.
In \textit{2023 IEEE 28th Pacific Rim International Symposium on Dependable Computing (PRDC)} (pp. 337--341). IEEE.

\bibitem{liu2024pcsp}
Liu, Z., Ma, M., Jiang, K., Hou, Z., Shi, L., \& Dong, J. S. (2024).
Pcsp\# denotational semantics with an application in sports analytics.
In \textit{The Application of Formal Methods: Essays Dedicated to Jim Woodcock on the Occasion of His Retirement} (pp. 71--102). Springer.

\bibitem{liu2024exploring}
Liu, Z., Dong, C., Wang, C., Dong, T. Y., \& Jiang, K. (2024).
Exploring team strategy dynamics in tennis doubles matches.
In \textit{International Sports Analytics Conference and Exhibition} (pp. 104--115). Springer.

\bibitem{liu2023sports}
Liu, Z., Jiang, K., \& Dong, J. S. (2023).
Sports injury prediction in professional tennis.
In \textit{2023 IEEE 28th Pacific Rim International Symposium on Dependable Computing (PRDC)} (pp. 304--308). IEEE.

\bibitem{liu2025analyzing}
Liu, Z., Dong, C., Chen, J. W., Jiang, A. M. J., Chen, G., Shaikh, A. F., Dong, T. Y., Wang, C., Jiang, K., \& Dong, J. S. (2025).
Analyzing the Formation Strategy in Tennis Doubles Game.
\textit{SN Computer Science}, 6(2), 100. Springer.

\bibitem{liu2024strategy}
Liu, Z., Durrani, M., Xuan, L. Y., Simon, J.-F., \& Deon, T. Y. F. (2024).
Strategy Analysis in NFL Using Probabilistic Reasoning.
In \textit{International Sports Analytics Conference and Exhibition} (pp. 116--128). Springer.

\bibitem{liu2025f}
Liu, Z., Jiang, K., Ma, M., Hou, Z., Lin, Y., \& Dong, J. S. (2025).
F$^3$Set: Towards Analyzing Fast, Frequent, and Fine-grained Events from Videos.
\textit{arXiv preprint arXiv:2504.08222}.

\bibitem{hundal2024soccer}
Hundal, R. S., Liu, Z., Wadhwa, B., Hou, Z., Jiang, K., \& Dong, J. S. (2024).
Soccer Strategy Analytics Using Probabilistic Model Checkers.
In \textit{International Sports Analytics Conference and Exhibition} (pp. 249--264). Springer.

\bibitem{jiang2020deep}
Jiang, K., Izadi, M., Liu, Z., \& Dong, J. S. (2020).
Deep learning application in broadcast tennis video annotation.
In \textit{2020 25th International Conference on Engineering of Complex Computer Systems (ICECCS)} (pp. 53--62). IEEE.

\bibitem{jiang2023court}
Jiang, K., Li, J., Liu, Z., \& Dong, C. (2023).
Court detection using masked perspective fields network.
In \textit{2023 IEEE 28th Pacific Rim International Symposium on Dependable Computing (PRDC)} (pp. 342--345). IEEE.

\bibitem{jiang2024tracking}
Jiang, K., Liu, Z., Wu, Q., Ma, M., \& Dong, J. S. (2024).
Tracking Small and Fast Moving Ball in Broadcast Videos Using Transfer Learning and the Enhanced Interactive Multi-motion Model.
In \textit{International Sports Analytics Conference and Exhibition} (pp. 81--96). Springer.


\end{thebibliography}
\end{document}